\documentclass[10pt,twocolumn,letterpaper]{article}

\usepackage{cvpr}

\usepackage{times}
\usepackage{epsfig}
\usepackage{graphicx}
\usepackage{amsmath}
\usepackage{amssymb}
\usepackage{color}
\usepackage{soul}
\usepackage{multirow}
\usepackage{arydshln}
\usepackage{amssymb}
\usepackage{pifont}
\usepackage{balance}
\makeatletter
\g@addto@macro\small{%
  \setlength\abovedisplayskip{-4pt}
  \setlength\abovedisplayshortskip{-4pt}
  \setlength\belowdisplayshortskip{-4pt}
  \setlength\belowdisplayskip{-4pt}
}
\makeatother

\newcommand{\para}[1]{\vspace{-13pt}\paragraph{#1}}

\cvprfinalcopy %

\def\cvprPaperID{****} %

\ifcvprfinal\fi

\newcommand{\eat}[1]{}

\newcommand{\dline}{\hdashline[0.5pt/1pt]}
\newcommand{\ddline}[1]{\cdashline{#1}[0.5pt/1pt]}

\newcommand{\stdevignore}[2]{${#1}$}

\newcommand{\nlstring}[1]{\emph{#1}}

\newcommand{\dataset}{\textsc{Touchdown}\xspace}
\newcommand{\spd}{SPD\xspace}
\newcommand{\tc}{TC\xspace}

\newcommand{\sed}{SED\xspace}

\newcommand{\navinstruction}{\bar{x}_n}
\newcommand{\sdrinstruction}{\bar{x}_s}

\newcommand{\instructionembed}{\textbf{x}}
\newcommand{\token}{x}

\newcommand{\orientation}{\alpha}

\newcommand{\states}{\mathcal{S}}
\newcommand{\state}{s}
\newcommand{\transition}{T}

\newcommand{\actions}{\mathcal{A}}
\newcommand{\action}{a}
\newcommand{\execution}{\bar{e}}

\newcommand{\act}[1]{{\tt\MakeUppercase{#1}}}
\newcommand{\stopaction}{\act{stop}}

\newcommand{\panorama}{\textbf{I}\xspace}
\newcommand{\imagefeature}{\textbf{F}\xspace}
\newcommand{\textimagefeature}{\textbf{G}\xspace}
\newcommand{\reconstructedfeature}{\textbf{H}\xspace}

\newcommand{\softmax}{\textsc{SoftMax}\xspace}
\newcommand{\embedding}{\varphi}

\newcommand{\bilstm}{{\rm BiLSTM}\xspace}
\newcommand{\hiddenstate}{\textbf{h}}
\newcommand{\textkernel}{\textbf{K}\xspace}
\newcommand{\convoperation}{\textsc{Conv}\xspace}
\newcommand{\deconvoperation}{\textsc{Deconv}\xspace}

\newcommand{\cmark}{\ding{51}}%
\newcommand{\xmark}{\ding{55}}%

\newcommand{\resnet}{{\textsc{ResNet18}}\xspace}
\newcommand{\cnn}{{\textsc{CNN}}\xspace}
\newcommand{\system}[1]{\textsc{#1}\xspace}
\newcommand{\imagetextconcat}{\system{Concat}}
\newcommand{\imagetextconcatconv}{\system{ConcatConv}}
\newcommand{\lingunet}{\system{LingUNet}}
\newcommand{\unet}{\system{UNet}}
\newcommand{\texttoconv}{\system{Text2Conv}}

\newcommand{\stopbaseline}{\textsc{Stop}\xspace}
\newcommand{\mostfrequent}{\textsc{Frequent}\xspace}
\newcommand{\randomwalk}{\textsc{Random}\xspace}
\newcommand{\gamodel}{\textsc{GA}\xspace}
\newcommand{\concatmodel}{\textsc{RConcat}\xspace}
\newcommand{\bcloning}{\textsc{Sup}\xspace}

\newcommand{\sdrrandom}{\textsc{Random}\xspace}
\newcommand{\sdraverage}{\textsc{Average}\xspace}
\newcommand{\sdrcenter}{\textsc{Center}\xspace}

\begin{document}

\title{\textsc{Touchdown}: Natural Language Navigation and Spatial Reasoning \\ in Visual Street Environments}

\author{Howard Chen\thanks{Work done at Cornell University.}\\
ASAPP Inc.\\
New York, NY\\
\hspace{-18pt}{\tt\small hchen@asapp.com} \\
\and
Alane Suhr \hspace{1em} Dipendra Misra \hspace{1em} Noah Snavely \hspace{1em} Yoav Artzi \\
Department of Computer Science \& Cornell Tech, Cornell University \\
New York, NY\\
{\tt\small \{suhr, dkm, snavely, yoav\}@cs.cornell.edu}
}

\maketitle
\begin{abstract}
We study the problem of jointly reasoning about language and vision through a navigation and spatial reasoning task. We introduce the \dataset task and dataset, where an agent must first follow navigation instructions in a real-life  visual urban environment, and then identify a location described in natural language to find a hidden object at the goal position. The data contains $9{,}326$ examples of English instructions and spatial descriptions paired with demonstrations. Empirical analysis shows the data presents an open challenge to existing methods, and qualitative linguistic analysis  shows that the data displays richer use of spatial reasoning compared to related resources.\footnote{The data is available at \href{https://github.com/lil-lab/touchdown}{https://github.com/lil-lab/touchdown}.}

\end{abstract}

\section{Introduction}
\label{sec:intro}

\begin{figure}[!ht]
\centering\footnotesize
\includegraphics[trim={0 119 438 9},clip,width=0.88\linewidth]{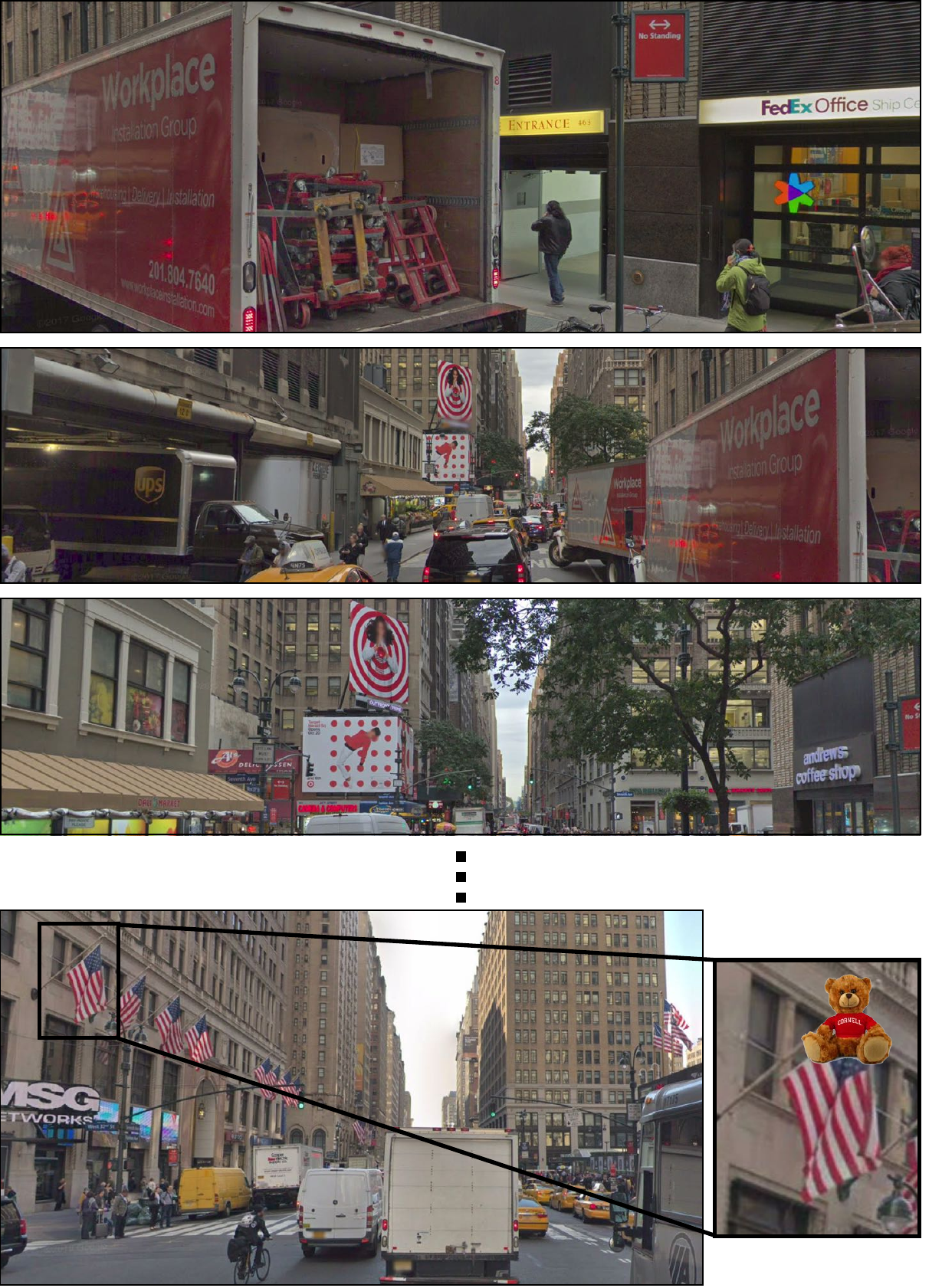}
\fbox{\begin{minipage}{0.95\linewidth}
\nlstring{Turn and go with the flow of traffic. At the first traffic light turn left. Go past the next two traffic light, As you come to the third traffic light you will see a white building on your left with many American flags on it. \ul{Touchdown is sitting in the stars of the first flag.}}
\end{minipage}}
\vspace{3pt}
\caption{An illustration of the task. The agent follows the instructions to reach the goal, starting by re-orientating itself (top image) and continuing by moving through the streets (two middle images). At the goal (bottom), the agent uses the spatial description (underlined) to locate Touchdown the bear. Touchdown only appears if the guess is correct (see bottom right detail).}
\vspace{-15pt}
\label{fig:intro}
\end{figure}

Consider the visual challenges of following natural language instructions in a busy urban environment. 
Figure~\ref{fig:intro} illustrates this problem. 
The agent must identify objects and their properties to resolve mentions to  \nlstring{traffic light}  and \nlstring{American flags},  identify patterns in how objects are arranged to find the \nlstring{flow of traffic},  and reason about how the relative position of objects changes as it moves to \nlstring{go past} objects. 
Reasoning about vision and language has been studied extensively with various tasks, including visual question answering~\cite{Antol:15vqa,Zitnick:13abstract}, visual navigation~\cite{Anderson:17,Misra:18goalprediction}, interactive question answering~\cite{Das:17eqa,gordon2018iqa}, and referring expression resolution~\cite{Kazemzadeh:14,Mao:16googleref,Matuszek:12}. 
However, existing work has largely focused on relatively simple visual input, including object-focused photographs~\cite{Lin:14coco,Reed:16} or simulated environments~\cite{Bisk:16dataset,Das:17eqa,Kolve:17,Misra:18goalprediction,Yan:18chalet}.
While this has enabled significant progress in visual understanding, the use of real-world visual input not only increases the  challenge of the vision task, it also drastically changes the kind of language it elicits and requires fundamentally  different reasoning.

In this paper, we study the problem of reasoning about vision and natural language using an interactive visual navigation environment based on Google Street View.\footnote{\href{https://developers.google.com/maps/documentation/streetview/intro}{https://developers.google.com/maps/documentation/streetview/intro}} 
We design the task of  first following  instructions to reach a goal position, and then  resolving a spatial description  at the goal by identifying the location in the observed image of Touchdown, a hidden teddy bear. 
Using this environment and task, we release \dataset,\footnote{
Touchdown is the \href{https://en.wikipedia.org/wiki/Touchdown_(mascot)}{unofficial mascot of Cornell University}.
} a dataset for navigation and spatial reasoning with real-life observations.

We design our task for diverse use of spatial reasoning, including for following instructions and resolving the spatial descriptions. 
Navigation requires the agent to reason about its relative position to objects and how these relations change as it moves through the environment. 
In contrast, understanding the description of the location of Touchdown requires the agent to reason about the spatial relations between observed objects. 
The two tasks also diverge in their learning challenges. While in both learning requires relying on indirect supervision to acquire spatial knowledge and language grounding, for navigation, the training data includes demonstrated actions, and for spatial description resolution, annotated target locations. 
The task can be addressed as a whole, or decomposed to its two portions. 


The key data collection challenge is designing a scalable process to obtain natural language data that reflects the richness of the visual input while discouraging overly verbose and unnatural language. 
In our data collection process, workers write and follow instructions. 
The writers navigate in the environment and hide Touchdown. 
Their goal is to make sure the follower can execute the instruction to find Touchdown. 
The measurable goal allows us to reward effective writers, and discourages overly verbose descriptions. 

We collect $9{,}326$ examples of the complete task, which decompose to the same number of navigation tasks and $27{,}575$ spatial description resolution (SDR) tasks. Each example is annotated with a navigation demonstration and the location of Touchdown. 
Our linguistically-driven  analysis shows the data requires significantly more complex reasoning than related datasets. 
Nearly all examples require resolving spatial relations between observable objects and between the agent and its surroundings, and each example contains on average $5.3$ commands and refers to $10.7$ unique entities in its environment. 


We empirically study the navigation and SDR tasks independently. 
For navigation, we focus on the performance of existing models trained with supervised learning. 
For SDR, we cast the problem of identifying Touchdown's location as an image feature reconstruction problem using a language-conditioned variant of the \unet architecture~\cite{DBLP:journals/corr/RonnebergerFB15,Misra:18goalprediction}. 
This approach  significantly outperforms several strong baselines.

\section{Related Work and Datasets}
\label{sec:related}


Jointly reasoning about vision and language has been studied extensively, most commonly focusing on static visual input for reasoning about image captions~\cite{Lin:14coco,Chen:15coco,Reed:16,Suhr:17visual-reason,Suhr2018nlvr2} and grounded question answering~\cite{Antol:15vqa,goyal2017making,Zitnick:13abstract}. 
Recently, the problem has been studied in interactive simulated  environments where the visual input changes as the agent acts, such as interactive question answering~\cite[]{Das:17eqa,gordon2018iqa} and instruction following~\cite{Misra:18goalprediction,Misra:17instructions}. 
In contrast, we focus on an interactive environment with real-world observations.


The most related resources to ours are R2R~\cite{Anderson:17} and Talk the Walk~\cite{devries:18}. 
R2R uses panorama graphs of house environments for the task of navigation instruction following. 
It includes $90$ unique environments, each containing an average of $119$ panoramas, significantly smaller than our  $29{,}641$ panoramas. 
Our larger environment requires following the instructions closely, as finding the goal using search strategies is unlikely, even given a large number of steps. 
We also observe that the language in our data is significantly more complex than in R2R (Section~\ref{sec:data-analysis}). 
Our environment setup is related to Talk the Walk, which uses panoramas in small urban environments for a navigation dialogue task. 
In contrast to our setup, the instructor does not observe the panoramas, but instead sees a simplified diagram of the environment with a small set of pre-selected landmarks. 
As a result, the instructor has less spatial information compared to \dataset. Instead the focus is on conversational coordination.


SDR is related to the task of referring expression resolution, for example as studied in ReferItGame~\cite{Kazemzadeh:14} and Google Refexp~\cite{Mao:16googleref}. 
Referring expressions describe an observed object, mostly requiring disambiguation between the described object and other  objects of the same type. 
In contrast, the goal of  SDR is to describe a specific location rather than discriminating. This leads to more complex language, as illustrated by the comparatively longer sentences of SDR (Section~\ref{sec:data-analysis}). 
Kitaev and Klein~\cite{kitaev2017misty} proposed a similar task to SDR, where given a spatial description and a small set of locations in a fully-observed simulated 3D environment, the system must select the location described from the set. 
We do not use distractor locations, requiring a system to consider all areas of the image to resolve a spatial description.

\section{Environment and Tasks}
\label{sec:task}

\begin{figure}[t]\centering
\includegraphics[trim={22 285 316 9},clip,width=0.75\linewidth]{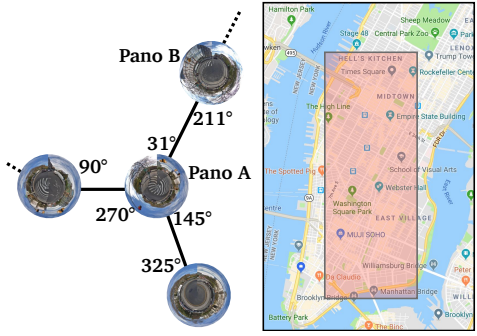}
\caption{An illustration of the environment. Left: part of the graph structure with polarly projected panoramas illustrating positions linked by edges, each labeled with its heading. Heading angles shown closer to each panorama represent the outgoing angle from that panorama; for example, the heading from Pano A to Pano B is $31^\circ$. Right: the area in New York City covered by the graph.}
\label{fig:env}
\vspace{-14pt}
\end{figure}

We use Google Street View to create a large navigation environment. Each position includes a 360$^\circ$  RGB panorama. 
The panoramas are connected in a graph-like structure with undirected edges connecting neighboring panoramas. Each edge connects to a panorama in a specific heading.  
For each panorama, we render perspective images for all headings that have edges. 
Our environment includes $29{,}641$ panoramas and $61{,}319$ edges from New York City. 
Figure~\ref{fig:env} illustrates the environment.


We design two tasks: navigation and spatial description resolution (SDR). 
Both tasks require recognizing objects and the spatial relations between them.
Navigation focuses on egocentric spatial reasoning, where instructions refer to the agent's relationship with its environment, including the objects it observes. 
The SDR task displays more allocentric reasoning, where the language requires understanding the relations between the observed objects to identify the target location. 
While navigation requires generating a sequence of actions from a small set of possible actions, 
SDR requires choosing a specific pixel in the observed image. 
Both tasks present different learning challenges. 
The navigation task could benefit from reward-based learning, while the SDR task  defines a supervised learning problem. 
The two tasks can be addressed separately, or combined by completing the SDR task at the goal position at the end of the navigation.


\subsection{Navigation}
\label{sec:task:nav}

The agent's goal is to follow a natural language instruction and reach a goal position. 
Let $\states$ be the set of all states. 
A state $\state \in \states$ is a pair $(\panorama, \orientation)$, where $\panorama$ is a panorama and $\orientation$ is the heading angle indicating the agent heading. 
We only allow states where there is an edge connecting to a neighboring panorama in the heading $\orientation$. 
Given a navigation instruction $\navinstruction$ and a start state $\state_1 \in \states$, the agent performs a sequence of actions. 
The set of actions $\actions$ is $\{ \act{forward},\act{left},\act{right},\stopaction\}$. 
Given a state $\state$ and an action $\action \in \actions$, the state is deterministically updated using a transition function $\transition : \states \times \actions \rightarrow  \states$. 
The $\act{forward}$ action moves the agent along the edge in its current heading. 
Formally, if the environment includes the edge $(\panorama_i, \panorama_j)$ at heading $\orientation$ in $\panorama_i$, the transition is $\transition((\panorama_i, \orientation), \act{forward}) = (\panorama_j, \orientation')$. 
The new heading $\orientation'$ is the heading of the edge  in $\panorama_j$ with the closest heading to $\orientation$. 
The $\act{left}$ ($\act{right}$) action changes the agent heading to the heading of the closest edge on the left (right). 
Formally, if the position panorama $\panorama$ has edges at headings $\orientation > \orientation' > \orientation''$, $\transition((\panorama, \orientation), \act{left}) = (\panorama, \orientation')$ and $\transition((\panorama, \orientation), \act{right}) = (\panorama, \orientation'')$. 
Given a start state $\state_1$ and a navigation instruction $\navinstruction$, an execution $\execution$ is a sequence of state-action pairs $\langle (\state_1, \action_1), ... , (\state_m, \action_m) \rangle$, where $\transition(\state_i, \action_i) = \state_{i+1}$ and $\action_m = \stopaction$. 




\para{Evaluation}
We use three evaluation metrics: task completion, shortest-path distance, and success-weighted edit distance. 
Task completion (\tc) measures the accuracy of completing the task  correctly. We consider an execution correct if the agent reaches the exact goal position or one of its neighboring nodes in the environment graph. 
Shortest-path distance (\spd) measures the mean distance in the graph between the agent's final panorama and the goal. \spd ignores turning actions and the agent heading. 
Success weighted by edit distance (\sed) is
  \mbox{$\frac{1}{N}\sum_{i=1}^{N} S_{i} (1 - \frac{{\rm lev}(\execution, \hat{\execution})}{\max(|\execution|, |\hat{\execution}|)})$},   
where the summation is over $N$ examples,  $S_i$ is a binary task completion indicator, $\execution$ is the reference execution, $\hat{\execution}$ is the predicted execution, ${\rm lev}(\cdot,\cdot)$ is the Levenshtein edit distance, and $|\cdot|$ is the execution length. 
The edit distance is normalized and inversed. 
We measure the distance and length over the sequence of panoramas in the execution, and ignore changes of orientation. \sed is related to success weighted by path length (SPL)~\cite{Anderson18:spl}, but is designed for instruction following in graph-based environments, where a specific correct path exists. 



\subsection{Spatial Description Resolution (SDR)}
\label{sec:task:sdr}

Given an image $\panorama$ and a natural language description $\sdrinstruction$, the task is to identify the point in the image that is referred to by the description. 
We instantiate this task as finding the location of Touchdown, a teddy bear, in the environment. Touchdown is hidden and not visible in the input. 
The image $\panorama$ is a 360$^\circ$ RGB panorama, and the output is a pair of $(x,y)$ coordinates specifying a location in the image. 

\para{Evaluation} We use three evaluation metrics: accuracy, consistency, and distance error. 
Accuracy is computed with regard to an annotated location. 
We consider a prediction as correct if the coordinates are within a slack radius of the annotation. We measure accuracy for radiuses of 40, 80, and 120 pixels and use  euclidean distance. 
Our data collection process results in multiple images for each sentence. We use this to measure consistency over unique sentences, which is measured similar to accuracy, but with a unique sentence considered correct only if all its examples are correct~\cite{Goldman:17}. We compute consistency for each slack value. 
We also measure the mean euclidean distance between the annotated location and the predicted location.


\section{Data Collection}
\label{sec:data-collection}

\newcommand{\tabimage}[2]{\begin{minipage}{0.315\textwidth}\footnotesize\centering {#1} \newline {#2} \end{minipage}}

\begin{figure*}
    \footnotesize
    \begin{tabular}{p{0.314\textwidth}p{0.314\textwidth}p{0.314\textwidth}}
    \multicolumn{3}{p{1.0\textwidth}}{\textbf{Task I: Instruction Writing} The worker starts at the beginning of the route facing north (a). The prescribed route is shown in the overhead map (bottom left of each image). The worker faces the correct direction and follows the path, while writing instructions that describe these actions (b). After following the path, the worker reaches the goal position, places Touchdown, and completes writing the instructions (c).} \\ 
    \multicolumn{3}{c}{\includegraphics[trim={14 342 153 85},clip,width=1.0\textwidth]{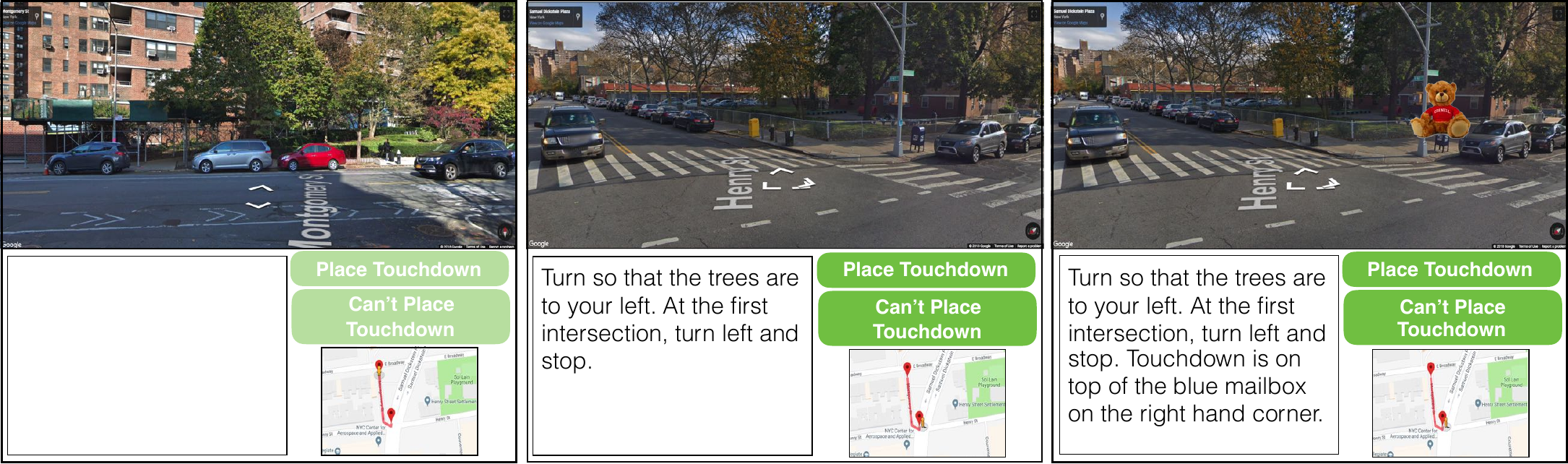}} \\
    \multicolumn{1}{c}{(a)} & \multicolumn{1}{c}{(b)} & \multicolumn{1}{c}{(c)} \\
     \multicolumn{3}{c}{\begin{minipage}{1.0\textwidth}\centering\vspace{0.3em}\includegraphics[trim={14 154 153 273},clip,width=1.0\textwidth]{content/figs/data_collection.pdf}\end{minipage}} \\   
    \textbf{Task II: Panorama Propagation} Given  the image from the leader's final position (top), including Touchdown's placement, and the instructions (right), the worker  annotates  the location of Touchdown in the neighboring image (bottom).  & \textbf{Task III: Validation} The  worker begins in the same heading as the leader, and follows the instructions  (bottom left) by navigating the environment. When the worker believes they have reached the goal, they  guess the target location by clicking in the Street View image. & \textbf{Task IV: Instruction Segmentation} The instructions are shown (left). The worker highlights segments corresponding to the navigation and target location subtasks. The highlighted segment is shown to the worker (right). 
    \end{tabular}
    \caption{Illustration of the data collection process.}
    \label{fig:data_collection}
    \vspace{-13pt}
\end{figure*}

%
%

We frame the data  collection process as a treasure-hunt task where a leader hides a treasure and writes directions to find it, and a follower follows the directions to find the treasure. 
The process is split into four crowdsourcing tasks (Figure~\ref{fig:data_collection}). 
The two main tasks are writing and following. 
In the writing task, a leader follows a prescribed route and hides Touchdown the bear at the end, while writing instructions that describe the path and how to find Touchdown.
The following task requires following the instructions from the same starting position to navigate and find Touchdown.
Additional tasks are used to segment the instructions into the navigation and target location tasks, and to propagate Touchdown's location to panoramas that neighbor the final panorama.
We use a customized Street View interface for data collection. 
However, the final data uses a static set of panoramas that do not require the Street View interface.

\para{Task I: Instruction Writing}
We generate routes by sampling start and end positions. 
The sampling process results in routes that often end in the middle of a city block. 
This encourages richer language, for example by requiring to describe the goal position rather than simply directing to the next intersection. The route generation details are described in the Supplementary Material. 
For each task, the worker is placed at the starting position facing north, and asked to follow a route specified in an overhead map view to a goal position. 
Throughout, they write instructions describing the path. 
The initial heading requires the worker to re-orient to the path, and thereby familiarize with their surroundings better. 
It also elicits interesting re-orientation instructions that often include references to the direction of objects (e.g., \nlstring{flow of traffic}) or their relation to the agent (e.g., \nlstring{the umbrellas are to the right}). 
At the goal panorama, the worker is asked to place Touchdown in a location of their choice that is not a moving object (e.g.,  a car or pedestrian) and to describe the location in their instructions.
The worker goal is to write instructions that a human follower can use to correctly navigate and locate the target without knowing the correct path or location of Touchdown. 
They are not permitted to write instructions that refer to text in the images, including street names, store names, or numbers.

\para{Task II: Target Propagation to Panoramas}
The writing task results in the location of Touchdown in a single panorama in the Street View interface. 
However, resolving the spatial description to the exact location is also possible from neighboring panoramas where the target location is visible. 
We use a crowdsourcing task to propagate the location of Touchdown to neighboring panoramas in the Street View interface, and to the identical panoramas in our static data. 
This  allows to complete the task correctly even if not stopping at the exact location, but still reaching a semantically equivalent position.  
The propagation in the Street View interface is used for our validation task. 
The task includes multiple steps. At each step, we show the instruction text and the original Street View panorama with Touchdown placed, and ask for the location for a single panorama, either from the Street View interface or from our static images. 
The worker can indicate if the target is occluded. 
The propagation annotation allows us to create multiple examples for each SDR, where each example uses the same SDR but shows the environment from a different position.


\para{Task III: Validation}
We use a separate task to validate each instruction. The worker is asked to follow the instruction in the customized Street View interface and find Touchdown. The worker sees only the Street View interface, and has no access to the overhead map. 
The task requires navigation and identifying the location of Touchdown. 
It is completed correctly if the  follower clicks within a $90$-pixel radius\footnote{This is roughly the size of Touchdown. The number is not directly comparable to the SDR accuracy measures due to different scaling.} of the ground truth target location of Touchdown. 
This requires the follower to be in the exact goal panorama, or in one of the neighboring panoramas we propagated the location to. 
The worker has five attempts to find Touchdown. Each attempt is a click. 
If the worker fails, we create another task for the same example to attempt again. 
If the second worker fails as well, the example is discarded.

\begin{figure}
\centering
\fbox{\begin{minipage}{0.45\textwidth} \footnotesize
\nlstring{\ul{Orient yourself in the direction of the red ladder. Go straight and take a left at the intersection with islands. Take another left at the intersection with a gray trash can to the left. Go straight until \textbf{near the end of the fenced in playground and court to the right near the end of the fenced in playground and court to the right.}}\textbf{ Touchdown is on the last basketball hoop to the right.}}
\end{minipage}}
\vspace{0.3em}
\caption{Example instruction where the annotated navigation (underlined) and SDR (bolded) segments overlap.}
\label{fig:seg_overlap}
\vspace{-10pt}
\end{figure}

\para{Task IV: Segmentation}
We annotate each token in the instruction to indicate if it describes the navigation or SDR tasks. This allows us to address the tasks separately. 
First, a worker highlights a consecutive prefix of tokens to indicate the navigation segment.
They then  highlight a suffix of tokens  for the SDR task. 
The navigation and target location segments may overlap (Figure~\ref{fig:seg_overlap}).



\begin{table}[!t]
\begin{footnotesize}
\begin{center}
\begin{tabular}{|l|c|} \hline
\multicolumn{1}{|c|}{\textbf{Task}} & \textbf{Number of Workers}  \\ \hline
Instruction Writing & $224$ \\
Target Propagation & $218$ \\
Validation & $291$ \\
Instruction Segmentation & $46$ \\ \hline
\end{tabular}
\end{center}
\end{footnotesize}
\vspace{-5pt}
\caption{Number of workers who participated in each task.}
\vspace{-10pt}
\label{tab:workers}
\end{table}

\para{Workers and Qualification}
We require passing a qualification task to do the writing task. 
The qualifier task requires correctly navigating and finding Touchdown for a predefined set of instructions. 
We consider workers that succeed in three out of the four tasks as qualified. 
The other three tasks do not require qualification.
Table~\ref{tab:workers} shows how many workers participated in each task.

\para{Payment and Incentive Structure}
The base pay for instruction writing is  $\$0.60$. 
For target propagation, validation, and segmentation we paid $\$0.15$,  $\$0.25$, and $\$0.12$.
We incentivize the instruction writers and followers with a bonus system. 
For each instruction that passes validation, we give the writer a bonus of $\$0.25$ and the follower a bonus of $\$0.10$. 
Both sides have an interest in completing the task correctly. 
The size of the graph makes it difficult, and even impossible, for the follower to complete the task and get the bonus if the instructions are wrong. 



\section{Data Statistics and Analysis}
\label{sec:data-analysis}

\begin{table}[t]
    \centering
    \footnotesize
    \begin{tabular}{|l|c|c|c|c|} \hline
       \multicolumn{1}{|c|}{\multirow{2}{*}{\textbf{Dataset}}} & \textbf{Dataset} & \textbf{Vocab.} & \textbf{Mean Text} & \textbf{Real} \\ 
       & \textbf{Size} & \textbf{Size} & \textbf{Length} & \textbf{Vision?} \\  \hline
        \dataset & $9{,}326$ & $5{,}625$ & $108.0$ & \multirow{3}{*}{\cmark} \\ 
        \hspace{0.3cm} Navigation & $9{,}326$ & $4{,}999$ & $89.6$ & \\
        \hspace{0.3cm} SDR & $25{,}575$ & $3{,}419$ & $29.7$ & \\ \ddline{1-5}
        \textsc{R2R}~\cite{Anderson:17} & $21{,}567$ & $3{,}156$ & $29.3$ & \cmark \\ \ddline{1-5}
        \textsc{SAIL}~\cite{MacMahon:06} & $706$ & $563$ & $36.7$ & \xmark \\ \ddline{1-5}
        \textsc{Lani}~\cite{Misra:18goalprediction} & $5{,}487$ & $2{,}292$ & $61.9$ & \xmark \\ \hline
    \end{tabular}
    \vspace{3pt}
    \caption{Data statistics of \dataset, compared to related corpora. For \dataset, we report statistics for the complete task, navigation only, and SDR only. Vocabulary size and text length are computed on the combined training and development sets.
    \textsc{SAIL} and \textsc{Lani} statistics  are computed using paragraph data.}
    \vspace{-10pt}
\label{tab:basic_stats}
\end{table}


\begin{figure}\centering
\begin{minipage}{0.45\textwidth}
\includegraphics[trim={130 320 120 320},clip,width=1.0\textwidth,page=1]{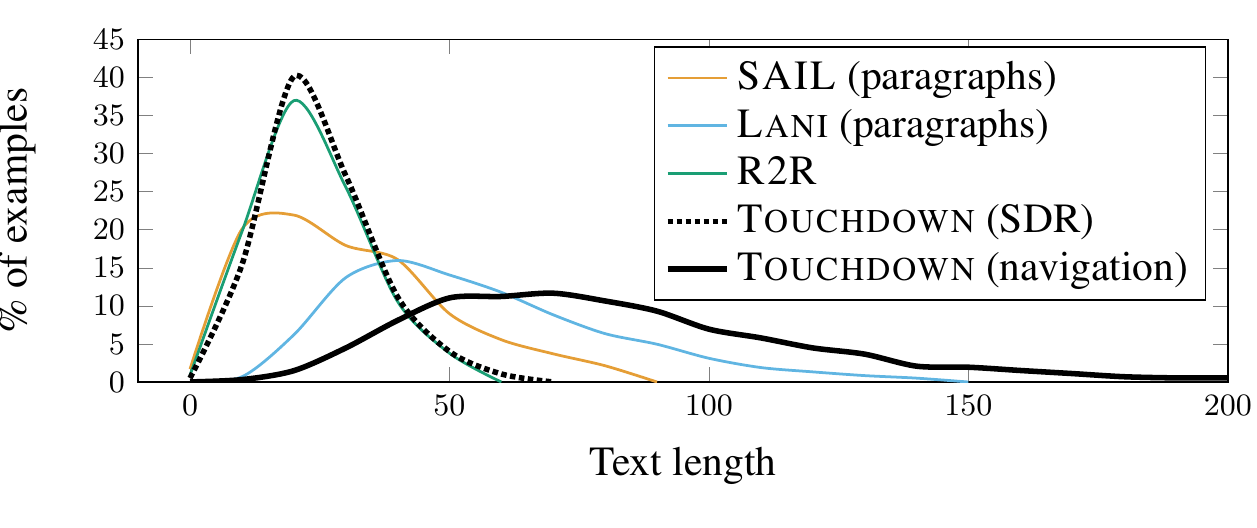}
\end{minipage}
\caption{Text lengths in \dataset and related corpora.}
\vspace{-5pt}
\label{fig:instruction_lengths}
\vspace{-10pt}
\end{figure}

Workers completed $11{,}019$ instruction-writing tasks, and $12{,}664$ validation tasks. 
$89.1\%$ examples were correctly validated, $80.1\%$ on the first attempt and $9.0\%$ on the second.\footnote{Several paths were discarded due to updates in Street View data.}
While we allowed five attempts at finding Touchdown during validation tasks, $64\%$ of the tasks  required a single attempt. The value of additional attempts decayed quickly: only $1.4\%$ of the tasks were only successful after five attempts. 
For the full task and navigation-only, \dataset includes $9{,}326$ examples  with $6{,}526$ in the training set, $1{,}391$ in the development set, and $1{,}409$ in the test set. For the SDR task, \dataset includes $9{,}326$ unique descriptions and $25{,}575$ examples with $17{,}880$ for training, $3{,}836$ for development, and $3{,}859$ for testing. 
We use our initial paths as gold-standard demonstrations, and the placement of Touchdown by the original writer as the reference location. 
Table~\ref{tab:basic_stats} shows basic data statistics. 
The mean instruction length is $108.0$ tokens. 
The average overlap between navigation and SDR is $11.4$ tokens.
Figure~\ref{fig:instruction_lengths} shows the distribution of text lengths.
Overall, \dataset contains a larger vocabulary and longer navigation instructions than  related corpora. 
The paths in \dataset are longer than in R2R~\cite{Anderson:17}, on average $35.2$ panoramas compared to $6.0$. 
SDR segments have a mean length of $29.8$ tokens, longer than in  common referring expression datasets; ReferItGame~\cite{Kazemzadeh:14} expressions $4.4$ tokens on average and Google RefExp~\cite{Mao:16googleref} expressions are $8.5$.


\begin{table*}[t]
\begin{footnotesize}
\begin{center}
\begin{tabular}{|l|c|c||c|c|c|c|c|c|p{7.8cm}|} 
	\hline
    \multicolumn{1}{|c|}{\multirow{3}{*}{\textbf{Phenomenon}}} &  \multicolumn{2}{c||}{\multirow{2}{*}{\textbf{\textsc{R2R}}}} & \multicolumn{6}{c|}{\textbf{\dataset}} & \multicolumn{1}{c|}{\multirow{3}{*}{\textbf{Example from \dataset}}} \\ \cline{4-9}
    \multicolumn{1}{|c|}{} & \multicolumn{2}{c||}{} & \multicolumn{2}{c|}{Overall} & \multicolumn{2}{c|}{Navigation} & \multicolumn{2}{c|}{SDR} & \multicolumn{1}{c|}{} \\ \cline{2-9}
    \multicolumn{1}{|c|}{}& $c$ & \stdevignore{\mu}{\sigma} & $c$ & \stdevignore{\mu}{\sigma} & $c$ & \stdevignore{\mu}{\sigma} & $c$ & \stdevignore{\mu}{\sigma} & \multicolumn{1}{c|}{}  \\
	\hline
	\hline
	Reference to  &  \multirow{2}{*}{$25$} &  \multirow{2}{*}{$3.7$} &  \multirow{2}{*}{$25$} &  \multirow{2}{*}{$10.7$} &  \multirow{2}{*}{$25$} &  \multirow{2}{*}{$9.2$} &  \multirow{2}{*}{$25$} &  \multirow{2}{*}{$3.2$} &  \multirow{2}{*}{\nlstring{\dots You'll pass \textbf{three trashcans} on your left \dots}} \\ 
	unique entity  & & & & & & & & & \\ \dline
	Coreference  & $8$ & $0.5$ & $22$ &$2.4$ &$15$& $1.1$ & $22$ & $1.5$ & \nlstring{\dots a brownish colored brick building with a black fence around \textbf{it}\dots } \\ \dline
	Comparison  & $1$ & $0.0$ & $6$ & $0.3$ & $3$ & $0.1$ & $5$ & $0.2$ & \nlstring{\dots The bear is in the middle of the \textbf{closest} tire.} \\ \dline
	Sequencing  & $4$ & $0.2$ & $22$& $1.9$& $21$ & $1.6$ &  $9$ & $0.4$ & \nlstring{\dots Turn left at the \textbf{next} intersection \dots} \\ \dline
	Count & $4$ & $0.2$ & $11$& $0.5$& $9$ & $0.4$ & $8$ & $0.3$ & \nlstring{\dots there are \textbf{two} tiny green signs you can see in the distance \dots} \\ \dline
	Allocentric  & \multirow{2}{*}{$5$} & \multirow{2}{*}{$0.2$} & \multirow{2}{*}{$25$} & \multirow{2}{*}{$2.9$} & \multirow{2}{*}{$17$} & \multirow{2}{*}{$1.2$} & \multirow{2}{*}{$25$} & \multirow{2}{*}{$2.2$} & \multirow{2}{*}{\nlstring{\dots There is a fire hydrant, the bear is \textbf{on top}}} \\ 
	spatial relation & & & & & & & & & \\ \dline
	Egocentric & \multirow{2}{*}{$20$} & \multirow{2}{*}{$1.2$} & \multirow{2}{*}{$25$} & \multirow{2}{*}{$4.0$} & \multirow{2}{*}{$23$} & \multirow{2}{*}{$3.6$} & \multirow{2}{*}{$19$} & \multirow{2}{*}{$1.1$} & \multirow{2}{*}{\nlstring{\dots up ahead there is some flag poles \textbf{on your right hand side}\dots }}\\ 
	spatial relation & & & & & & & & & \\ \dline
	Imperative & $25$ & $4.0$ & $25$& $5.3$& $25$ & $5.2$ & $4$ & $0.2$ & \nlstring{\dots \textbf{Enter} the next intersection and \textbf{stop} \dots}\\ \dline
	Direction & $22$ & $2.8$ & $24$& $3.7$& $24$ & $3.7$ & $1$ & $0.0$ & \nlstring{\dots Turn 
	\textbf{left}. Continue \textbf{forward} \dots}  \\ \dline
	Temporal condition  & $7$ & $0.4$ & $21$& $1.9$& $21$ & $1.9$ & $2$ & $0.1$ & \nlstring{\dots Follow the road \textbf{until you see} a school on your right\dots } \\ \dline
	State verification  & $2$ & $0.1$ & $21$& $1.8$& $18$ & $1.5$ & $16$ & $0.8$ & \nlstring{\dots \textbf{You should see} a small bridge ahead \dots}\\ \hline
\end{tabular}
\end{center}
\end{footnotesize}
\vspace{-5pt}
\caption{Linguistic analysis of $25$ randomly sampled development examples in \dataset and \textsc{R2R}. 
We annotate each example for the presence and count of each phenomenon.
We distinguish statistics for the entire text, navigation, and SDR segments in \dataset.
$c$ is the number of instructions out of the $25$ containing at least one example of the phenomenon; $\mu$ is the mean number of times each phenomenon appears in each of the $25$ instructions.}
\vspace{-15pt}
\label{tab:ling}
\end{table*}

We perform qualitative linguistic analysis of \dataset  to understand the type of reasoning required to solve the navigation and SDR tasks. 
We identify a set of phenomena, and randomly sample $25$ examples from the development set, annotating each with the number of times each phenomenon occurs in the text.
Table~\ref{tab:ling} shows results comparing \dataset with R2R.\footnote{See the Supplementary Material for analysis of \textsc{SAIL} and \textsc{Lani}.}
Sentences in \dataset refer to many more unique, observable entities ($10.7$ vs $3.7$), and almost all examples in \dataset include coreference to a previously-mentioned entity.
More examples in \dataset require reasoning about counts, sequences, comparisons, and spatial relationships of objects.
Correct execution in \dataset requires taking actions only when certain conditions are met, and ensuring that the agent's observations match a described scene, while this is rarely required in R2R.
Our data is rich in spatial reasoning. We distinguish two types: between multiple objects (\emph{allocentric}) and between the agent and its environment (\emph{egocentric}).
We find that navigation segments contain more egocentric spatial relations than SDR segments, and SDR segments require more allocentric reasoning.
This corresponds to the two tasks: navigation mainly requires moving the agent relative to its environment, while SDR requires resolving a point in space relative to other objects.






\section{Spatial Reasoning with \lingunet}

We cast the SDR task as a language-conditioned image reconstruction problem, where we predict a distribution of the location of Touchdown over the entire observed image. 

\subsection{Model}

We use the \lingunet architecture~\cite{Misra:18goalprediction,Blukis:18visit-predict}, which was originally introduced for goal prediction and planning in instruction following. 
\lingunet is a language-conditioned variant of the \unet architecture~\cite{DBLP:journals/corr/RonnebergerFB15}, an image-to-image encoder-decoder architecture  widely used for image segmentation. 
\lingunet incorporates language into the image reconstruction phase to fuse the two modalities.
We modify the architecture to predict a probability distribution over the input panorama image. 



We process the description text tokens $\sdrinstruction = \langle \token_1, \token_2, \dots ,\token_l \rangle$ using a bi-directional Long Short-term Memory (LSTM) recurrent neural network to generate $l$ hidden states. The forward computation is $\hiddenstate_{i}^f = \bilstm( \embedding ( \token_{i} ) ,\hiddenstate_{i - 1}^f), i = 1, \dots, l$, where $\embedding$ is a learned word embedding function. We compute the backward hidden states $\hiddenstate_{i}^b$ similarly. The text representation is an average of the concatenated hidden states $\instructionembed = \frac{1}{l}\sum_{i=1}^l [h^f_{i}; h^b_{i}]$. 
We map the RGB panorama $\panorama$ to a feature representation $\imagefeature_0$ with a pre-trained \resnet~\cite{He:16resnet}. 



\begin{figure}[t]\centering
\includegraphics[trim={0 250 430 175},clip,width=1.0\linewidth]{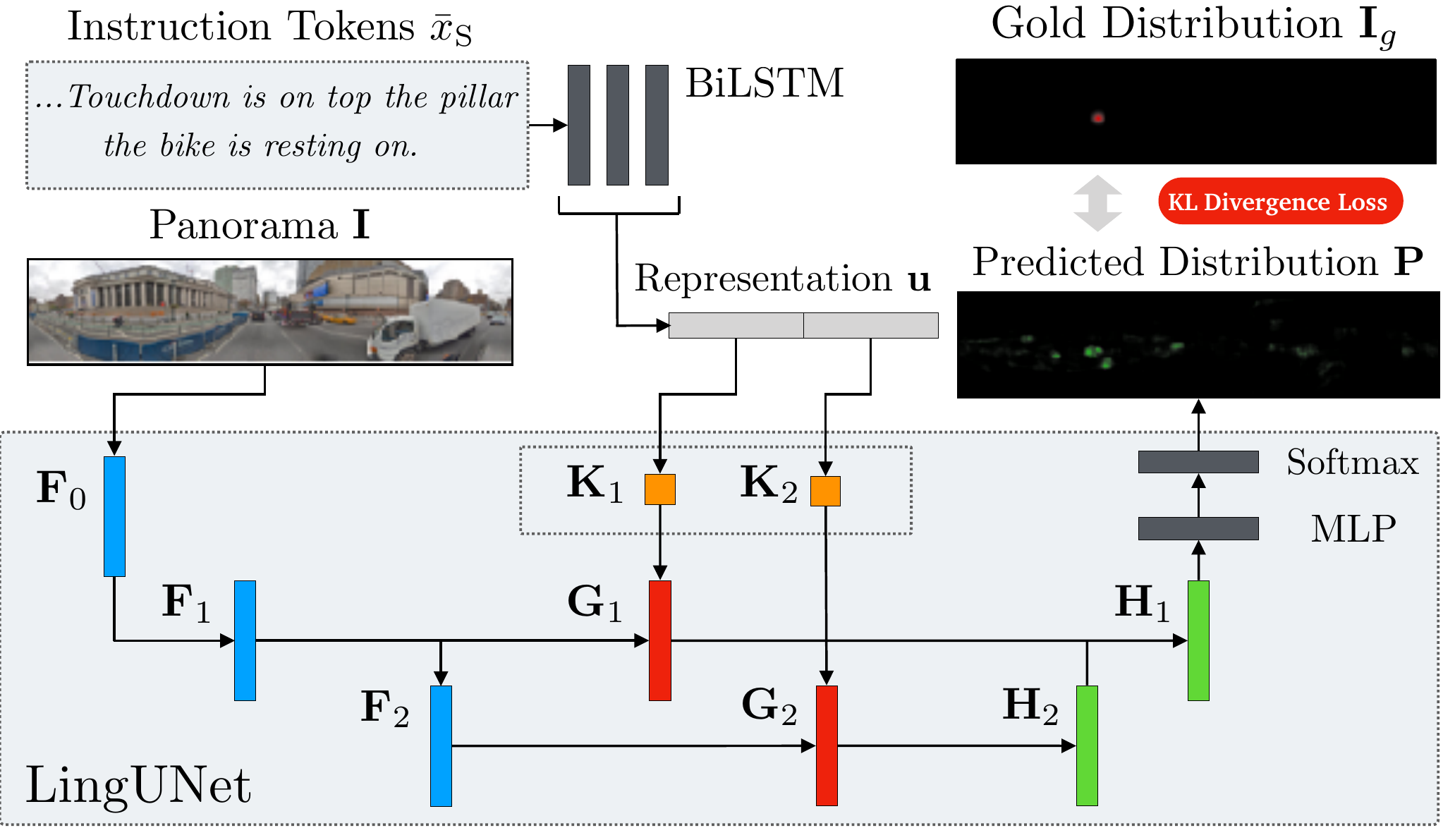}
\vspace{-12pt}
\caption{LingUNet architecture with two layers ($m=2$).}
\vspace{-12pt}
\label{fig:lingunet.pdf}
\end{figure}


\lingunet performs $m$ levels of convolution and deconvolution operations. 
We generate a sequence of feature maps $\imagefeature_{k} = \cnn_{k}(\imagefeature_{k - 1}), k = 1, \dots ,m$ with learned convolutional layers $\cnn_k$. 
We slice the text representation $\instructionembed$ to $m$ equal-sized slices, and reshape each with a linear projection to a $1 \times 1$ filter  $\textkernel_{k}$. 
We convolve each feature map $\imagefeature_{k}$ with $\textkernel_{k}$ to obtain a text-conditioned feature map $\textimagefeature_{k} = \convoperation(\textkernel_{k}, \imagefeature_{k})$. 
We use $m$ deconvolution operations to generate feature maps of increasing size to create $\reconstructedfeature_{1}$:

\begin{small}
\begin{equation}
\nonumber  \reconstructedfeature_{k} =   \begin{cases}
    \deconvoperation_{k}([\reconstructedfeature_{k+1}; \textimagefeature_{k}]), & \text{if $k=1, \dots, m - 1$}\\
    \deconvoperation_{k}(\textimagefeature_{k}), & \text{if $k=m$}\;\;.
  \end{cases}
\end{equation}
\end{small}

\noindent
We compute a single value for each pixel by projecting  the channel vector for each pixel using a single-layer perceptron with a ReLU non-linearity. Finally, we compute a probability distribution over the feature map using a \softmax. 
The predicted location is the mode of the distribution. 

\subsection{Experimental Setup}

The evaluation metrics are described in Section~\ref{sec:task:sdr} and the data in Section~\ref{sec:data-analysis}. 

\para{Learning}
We use supervised learning. The gold label is a Gaussian smoothed distribution. The coordinate of the maximal value of the distribution is the exact coordinate where Touchdown is placed. We minimize the KL-divergence between the Gaussian and the predicted distribution. 




\para{Systems} 
We evaluate three non-learning baselines: 
(a) \system{Random}: predict a pixel at random; 
(b) \system{Center}: predict the center pixel;
(c) \system{Average}: predict the average pixel, computed over the training set. 
In addition to a two-level \lingunet ($m=2$), we  evaluate three learning baselines: \imagetextconcat, \imagetextconcatconv, and \texttoconv. The first two compute a \resnet feature map representation of the image and then fuse it with the text representation to compute pixel probabilities. The third uses the text  to compute kernels to convolve over the \resnet image representation. 
The Supplementary Material provides further details.

\subsection{Results}

Table~\ref{tab:sdr-results} shows development and test results. 
The low performance of the non-learning baselines illustrates the challenge of the task. 
We also experiment with a \unet architecture that is similar to our \lingunet but has no access to the language. 
This result illustrates that visual biases exist in the data, but only enable relatively low performance. 
All the learning systems outperform the non-learning baselines and the \unet, with \lingunet performing best. 

\begin{figure*}[t]\centering
\frame{\includegraphics[trim={0 620 290 30},clip,width=0.83\linewidth]{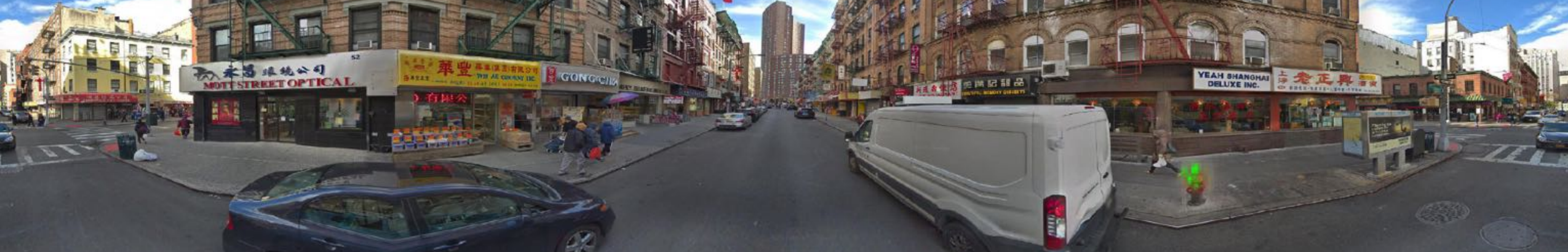}} \\[2pt]
\fbox{\begin{minipage}{0.95\linewidth}
\nlstring{there will be a white/grey van parked on the right side of the road, and right behind the van on the walkway, there is a black fire hydrant with silver top, the touchdown is on the silver part of the fire hydrant.}
\end{minipage}}\\[2pt]
\frame{\includegraphics[trim={0 470 290 180},clip,width=0.83\linewidth]{content/figs/sdr_example.pdf}} \\[2pt]
\fbox{\begin{minipage}{0.95\linewidth}
\nlstring{a black doorway with red brick to the right of it, and green brick to the left of it. it has a light just above the doorway, and on that light is where you will find touchdown.}
\end{minipage}}
\vspace{2pt}
\caption{SDR pixel-level predictions with \lingunet. Red-overlaid pixels indicate the Gaussian smoothed target location. Bright green overlay indicates the model's predicted probability distribution over pixels.}
\vspace{-12pt}
\label{fig:sdr-prediction}
\end{figure*}

Figure~\ref{fig:sdr-prediction} shows  pixel-level predictions  using \textsc{LingUNet}. 
The distribution prediction is visualized as a heatmap overlaid on the image. 
\lingunet often successfully solves descriptions anchored in objects that are unique in the image, such the \nlstring{fire hydrant} at the top image. 
The lower example is more challenging. While the model correctly reasons that Touchdown is on \nlstring{a light just above the doorway}, it fails to find the exact door. Instead,  the probability distribution is shared between multiple similar locations, the space above three other doors in the image.

\newcommand{\consistency}[1]{/ #1}

\begin{table}[t]
    \footnotesize
    \centering
    \bgroup
    \setlength{\tabcolsep}{4pt}
    \begin{tabular}{|l|c|c|c|c|}
    \hline
    \multicolumn{1}{|c|}{\textbf{Method}} & \textbf{A/C@40px} & \textbf{A/C@80px} & \textbf{A/C@120px} & \textbf{Dist} \\
    \hline\hline
    \multicolumn{5}{|l|}{\textbf{Development Results}} \\
    \hline \hline
        {\sdrrandom} & $0.18$ \consistency{$0.00$} & $0.59$ \consistency{$0.00$} & $1.28$  \consistency{$0.00$}& $1185$ \\
        {\sdrcenter} & $0.55$ \consistency{$0.07$}& $1.62$ \consistency{$0.07$}& $3.26$ \consistency{$0.36$}& $777$ \\
        {\sdraverage} & $1.88$ \consistency{$0.07$}& $4.22$ \consistency{$0.29$}& $7.14$ \consistency{$0.79$}& $762$ \\
        \hline
        {\unet} & $10.86$ \consistency{$2.69$}& $13.94$ \consistency{$3.31$}& $16.69$ \consistency{$3.91$}& $957$ \\
    \hline
        {\imagetextconcat} &  $13.70$ \consistency{$3.22$}& $17.85$ \consistency{$4.46$}& $21.16$ \consistency{$5.47$} & $917$ \\
        {\imagetextconcatconv} & $13.56$ \consistency{$3.24$}& $18.00$ \consistency{$4.58$}& $21.42$ \consistency{$5.71$}& $918$ \\
        {\texttoconv} & $24.03$ \consistency{$7.60$}& $29.36$ \consistency{$10.02$}& $32.60$ \consistency{$11.42$}& $783$ \\
        {\lingunet} & \textbf{$24.81$} \consistency{\textbf{$7.73$}}& \textbf{$32.83$} \consistency{\textbf{$13.00$}}& \textbf{$36.44$} \consistency{\textbf{$15.01$}}& \textbf{$729$} \\
        \hline \hline
    \multicolumn{5}{|l|}{\textbf{Test Results}} \\
    \hline\hline
        {\sdrrandom} & $0.21$ \consistency{$0.00$}& $0.78$ \consistency{$0.00$}& $1.89$ \consistency{$0.00$}& $1179$ \\
        {\sdrcenter} & $0.31$ \consistency{$0.00$}& $1.61$ \consistency{$0.21$}& $3.93$ \consistency{$0.57$}& $759$ \\
        {\sdraverage} & $2.43$ \consistency{$0.07$}& $5.21$ \consistency{$0.57$}& $7.96$ \consistency{$1.06$}& $744$ \\
        \hline
        {\texttoconv} & $24.82$ \consistency{$8.21$}& $30.40$ \consistency{$11.73$}& $34.13$ \consistency{$13.32$}& $747$ \\
        {\lingunet} & \textbf{$26.11$} \consistency{\textbf{$8.80$}}& \textbf{$34.59$} \consistency{\textbf{$14.57$}}& \textbf{$37.81$} \consistency{\textbf{$16.11$}}& \textbf{$708$} \\
    \hline
    \end{tabular}
    \egroup
    \vspace{2pt}
    \caption{Development and test results on the SDR task. We report accuracy/consistency (A/C) with different thresholds ($40$, $80$, and $120$) and mean distance error.}
    \vspace{-15pt}
    \label{tab:sdr-results}
\end{table}

\section{Navigation Baselines}



\subsection{Methods and Setup} 

We evaluate three non-learning baselines: (a) \stopbaseline: agent stops immediately; (b) \randomwalk: Agent samples non-stop actions uniformly until reaching the action horizon; and (c) \mostfrequent: agent always takes the most frequent action in the training set (\act{forward}). 
We also evaluate two recent navigation models: (a) \gamodel: gated-attention~\cite{chaplot2017gated}; and (b) \concatmodel: a recently introduced model for landmark-based navigation in an environment that uses Street View images~\cite{mirowski2018learning}. 
We represent the input images with \resnet features similar to the SDR task. 

We  use   asynchronous training using multiple clients to generate rollouts on different partitions of the training data. We compute the gradients and updates using \textsc{Hogwild!}~\cite{recht2011hogwild} and \textsc{Adam} learning rates~\cite{Kingma:14adam}. 
We use supervised learning by maximizing the log-likelihood of actions in the reference demonstrations.


The details of the models, learning, and hyperparameters are provided in the Supplementary Material.


\subsection{Results}

Table~\ref{tab:dev-nav-results} shows development and test results for our three valuation metrics (Section~\ref{sec:task:nav}). The \stopbaseline, \mostfrequent and \randomwalk illustrate the complexity of the task. The learned baselines perform better. 
We observe that \concatmodel outperforms \gamodel across all three metrics. 
In general though, the performance illustrates the challenge of the task. 
Appendix~\ref{sec:app:nav} includes additional navigation experiments, including single-modality baselines. 

\section{Complete Task Performance}

We use a simple pipeline combination of the best models of the SDR and navigation tasks to complete the full task. 
Task completion is measured as finding Touchdown. We observe an accuracy of $4.5$\% for a threshold of $80$px. 
In contrast, human performance is significantly higher. 
We estimate human performance using our annotation statistics~\cite{Suhr2018nlvr2}. 
To avoid spam and impossible examples, we consider only examples that were successfully validated. We then measure the performance of workers that completed over $30$ tasks for these valid examples. This includes $55$ workers. Because some examples required multiple tries to validate this set includes tasks that workers failed to execute but were later validated. 
The mean performance across this set of workers using the set of valid tasks is $92\%$ accuracy.





\begin{table}[t]
  \footnotesize
    \centering
    \begin{tabular}{|l|c|c|c|}
    \hline
    \multicolumn{1}{|c|}{\textbf{Method}} & \textbf{TC} & \textbf{SPD} & \textbf{SED} \\
    \hline\hline
    \multicolumn{4}{|l|}{\textbf{Development Results}} \\
    \hline\hline
        \stopbaseline &  $0.0$ & $26.7$ & $0.0\phantom{00}$\\
        \mostfrequent & $0.1$ & $52.3$ & $0.001$\\
        \randomwalk & $0.2$ & $26.8$ & $0.001$\\
        \hline 
        \gamodel  & $7.9$ & $21.5$ & $0.077$ \\
        \concatmodel  & $\textbf{9.8}$ & $\textbf{19.1}$ & $\textbf{0.094}$\\
        \hline\hline
    \multicolumn{4}{|l|}{\textbf{Test Results}} \\
    \hline\hline
        \stopbaseline & $0.0$ & $27.0$ & $0.0\phantom{00}$ \\
        \mostfrequent & $0.0$ & $53.1$ & $0.0\phantom{00}$\\
        \randomwalk & $0.2$ & $26.9$ & $0.001$\\
        \hline 
        \gamodel & $5.5$ & $21.3$ & $0.054$\\
        \concatmodel  & $\textbf{10.7}$ & $\textbf{19.5}$ & $\textbf{0.104}$\\     
        \hline
    \end{tabular}
    \vspace{2pt}
    \caption{Development and test  navigation results.}
    \label{tab:dev-nav-results}
    \vspace{-15pt}
\end{table}

\section{Data Distribution and Licensing}
\label{sec:dist}

We release the environment graph as panorama IDs and edges, scripts to download the RGB panoramas using the  Google API, the collected data, and our code at \href{http://touchdown.ai}{touchdown.ai}.  
These parts of the data are released with a CC-BY $4.0$ license. 
Retention of downloaded panoramas should follow Google's policies. 
We also release \resnet image features of the RGB panoramas through a request form. 
The complete license is available with the data.


%

%

%

\vspace{-3pt}
\section{Conclusion}
\label{sec:disc}
\vspace{-3pt}

We introduce \dataset, a dataset for natural language navigation and spatial reasoning using real-life visual observations. 
We define two tasks that require addressing a diverse set of reasoning and learning challenges. 
Our linguistically-driven analysis shows the data presents complex spatial reasoning challenges. 
This illustrates the benefit of using visual input that reflects the type of observations people see in their daily life, and demonstrates the effectiveness of our goal-driven data collection process. 



\vspace{-5pt}
\section*{Acknowledgements}

This research was supported by a Google Faculty Award, NSF award CAREER-1750499, NSF Graduate Research Fellowship DGE-1650441, and the generosity of Eric and Wendy Schmidt by recommendation of the Schmidt Futures program. 
We wish to thank Jason Baldridge for his extensive help and advice, and Valts Blukis and the anonymous reviewers for their helpful comments.

\balance

{\small
\bibliographystyle{ieee}
\bibliography{main}
}

\clearpage
\appendix

\section{Route Generation}

We generate each route by randomly sampling two panoramas in the environment graph, and querying the Google Direction API\footnote{\href{https://developers.google.com/maps/documentation/directions/start}{https://developers.google.com/maps/documentation/directions/start}} to obtain a route between them that follows correct road directions. 
Although the routes follow the direction of allowed traffic, the panoramas might still show moving against the traffic in two-way streets depending on which lane was used for the original panorama collection by Google. 
The route is segmented into multiple routes with length sampled uniformly between $35$ and $45$. 
We do not discard the suffix route segment, which may be shorter. 
Some final routes had gaps due to our use of the API. 
If the number of gaps is below three, we heuristically connect the detached parts of the route by adding intermediate panoramas, otherwise we remove the route segment.
Each of the route segments is used in a separate instruction-writing task. Because panoramas and route segments are sampled randomly, the majority of route segments stop in the middle of a block, rather than at an intersection.
This explicit design decision requires instruction-writers to describe exactly where in the block the follower should stop, which elicits references to a variety of object types, rather than simply referring to the location of an intersection.

\section{Additional Data Analysis}

\begin{table*}[t]
\begin{footnotesize}
\begin{center}
\begin{tabular}{|l|c|c|c|c||c|c|c|c|c|c|} 
	\hline
    \multicolumn{1}{|c|}{\multirow{3}{*}{ \textbf{Phenomenon}}} & \multicolumn{2}{|c|}{\textbf{\textsc{SAIL}}~\cite{MacMahon:06}} & \multicolumn{2}{|c||}{\textbf{\textsc{Lani}}~\cite{Misra:18goalprediction}} & \multicolumn{6}{|c|}{\textbf{\dataset}} \\ \cline{6-11}
    \multicolumn{1}{|c|}{} & \multicolumn{2}{|c|}{Paragraphs} & \multicolumn{2}{|c||}{Paragraphs} &  \multicolumn{2}{|c|}{Overall} & \multicolumn{2}{|c|}{Navigation} & \multicolumn{2}{|c|}{SDR}  \\ \cline{2-11}
     \multicolumn{1}{|c|}{} & $c$ & \stdevignore{\mu}{\sigma} & $c$ & \stdevignore{\mu}{\sigma} & $c$ & \stdevignore{\mu}{\sigma} & $c$ & \stdevignore{\mu}{\sigma}& $c$ & \stdevignore{\mu}{\sigma}\\
	\hline
	\hline
    Reference to unique entity  & $24$ & $4.0$ & $25$ & $7.2$ & $25$ & $10.7$ & $25$ & $9.2$ & $25$ & $3.2$ \\ \dline
	Coreference  & $12$ & $0.6$ & $22$ & $2.9$ & $22$ & $2.4$ &$15$& $1.1$ & $22$ & $1.5$ \\ \dline
    Comparison   & $0$ & $0.0$ & $2$ & $0.1$ & $6$ & $0.3$ & $3$ & $0.1$ & $5$ & $0.2$\\ \dline
    Sequencing  & $4$ & $0.2$  & $2$ & $0.1$ & $22$ & $1.9$& $21$ & $1.6$ &  $9$ & $0.4$ \\ \dline
	Count  & $16$ & $1.7$ & $2$ & $0.1$ & $11$ & $0.5$ & $9$ & $0.4$ & $8$ & $0.3$ \\ \dline
    Allocentric spatial relation  & $9$ & $0.4$ & $3$ & $0.2$ & $25$ & $2.9$ & $17$ & $1.2$ & $25$ & $2.2$ \\ \dline
    Egocentric spatial relation  & $13$ & $0.8$ & $24$ & $4.1$ & $25$ & $4.0$ & $23$ & $3.6$ & $19$ & $1.1$ \\ \dline
    Imperative & $23$ & $4.5$ & $25$ & $9.0$ & $25$ & $5.3$ & $25$ & $5.2$ & $4$ & $0.2$ \\ \dline
    Direction  & $23$ & $4.5$ & $25$ & $5.8$ & $24$ & $3.7$ & $24$ & $3.7$ & $1$ & $0.0$ \\ \dline
    Temporal condition   & $14$ & $0.7$ &$19$ & $2.0$ &$21$ & $1.9$ & $21$ & $1.9$ & $2$ & $0.1$ \\ \dline
    State verification   & $11$ & $0.5$ & $0$ & $0.0$ & $21$ & $1.8$ & $18$ & $1.5$ & $16$ & $0.8$ \\ \hline
\end{tabular}
\end{center}
\end{footnotesize}
\caption{Linguistic analysis of $25$ randomly sampled development examples in \dataset, \textsc{SAIL}, and \textsc{Lani}. 
}
\label{tab:cats_lanichai}
\end{table*}

\begin{figure*}[hbt!]\centering
\includegraphics[trim={0 120 660 0},clip,width=0.7\linewidth]{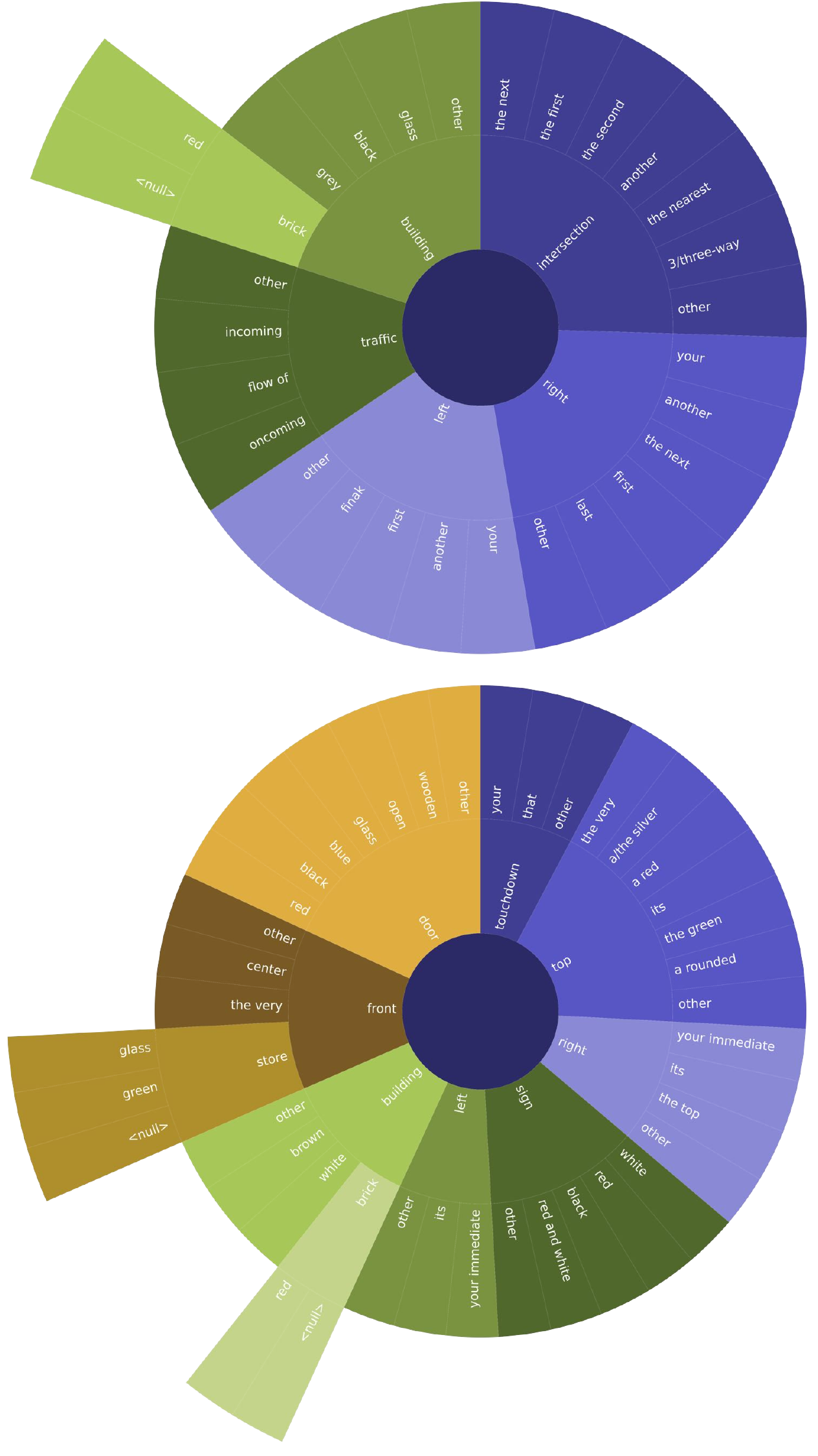}
\caption{An illustration of the referential language in our navigation (top) and SDR (bottom) instructions. We ranked all nouns by frequency and removed stop words. We show the top five/eight nouns (most inner circle) for navigation and SDR. For each noun, we show the most common modifiers that prefix it. The size of each segment is not relative to the frequency in the data.}
\label{fig:pie}
\end{figure*}

We perform linguistically-driven analysis to two additional navigation datasets: SAIL~\cite{MacMahon:06,Chen:11} and LANI~\cite{Misra:18goalprediction}, both using simulated environments. 
Both datasets include paragraphs segmented into single instructions. We performs our analysis at the paragraph level. 
We use the same categories as in Section~\ref{sec:data-analysis}. 
Table~\ref{tab:cats_lanichai} shows the analysis results. 
In general, in addition to the more complex visual input, \dataset displays similar or increased linguistic diversity compared to \textsc{Lani} and \textsc{SAIL}. 
\textsc{Lani} contains a similar amount of coreference, egocentric spatial relations, and temporal conditions, and more examples than \dataset of imperatives and directions.
\textsc{SAIL} contains a similar number of imperatives, and more examples of counts than \dataset. 
We also visualize some of the common nouns and modifiers observed in our data (Figure~\ref{fig:pie}).


\section{SDR Pixel-level Predictions}
\begin{figure*}[t!]
\centering
\fbox{\begin{minipage}{0.98\linewidth}
\nlstring{the dumpster has a blue tarp draped over the end closest to you. touchdown is on the top of the blue tarp on the dumpster.}
\end{minipage}} \\[10pt]
\begin{minipage}{0.98\linewidth}
\noindent
\textbf{\lingunet} The model correctly predicts the location of Touchdown, putting most of the predicted distribution (green) on the top-left of the dumpster at the center. 
\end{minipage} 
\frame{\includegraphics[trim={0 615 290 0},clip,width=0.98\linewidth]{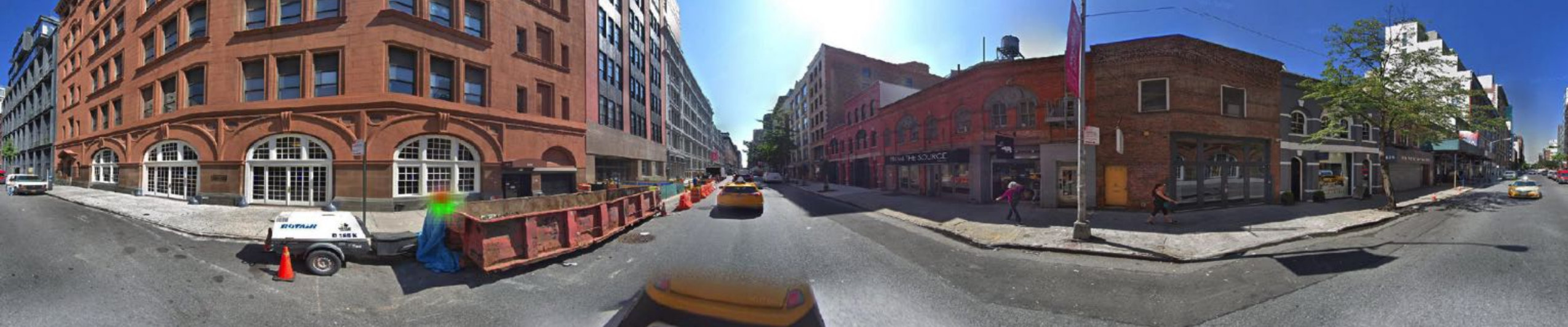}} \\[10pt]
\begin{minipage}{0.98\linewidth}
\noindent
\textbf{\texttoconv} The model incorrectly predicts the location of Touchdown to the top of the car on the far right. While some of the probability mass is correctly placed on  the dumpster, the pixel with the highest probability is on the car.
\end{minipage} 
\frame{\includegraphics[trim={0 445 290 170},clip,width=0.98\linewidth]{content/figs/sdr_sup_examples.pdf}} \\[10pt]
\begin{minipage}{0.98\linewidth}
\noindent
\textbf{\imagetextconcatconv} The model correctly predicts the location of Touchdown. The distribution is heavily concentrated at a couple of nearby pixels. 
\end{minipage} 
\frame{\includegraphics[trim={0 280 290 340},clip,width=0.98\linewidth]{content/figs/sdr_sup_examples.pdf}}\\[10pt]
\begin{minipage}{0.98\linewidth}
\noindent
\textbf{\imagetextconcat} The prediction is similar to \imagetextconcatconv.
\end{minipage} 
\frame{\includegraphics[trim={0 105 290 510},clip,width=0.98\linewidth]{content/figs/sdr_sup_examples.pdf}}
\caption{Three of the models are  doing fairly well. Only  \texttoconv fails to predict the location of Touchdown.}
\label{fig:sdr_sup_exp_1}
\end{figure*}

\begin{figure*}[t!]\centering
\fbox{\begin{minipage}{0.98\linewidth}
\nlstring{turn to your right and you will see a green trash barrel between the two blue benches on the right. click to the base of the green trash barrel to find touchdown.}
\end{minipage}} \\[10pt]
\begin{minipage}{0.98\linewidth}
\noindent
\textbf{\lingunet} The model accurately predicts the green trash barrel on the right as Touchdown's location.
\end{minipage} 
\frame{\includegraphics[trim={0 615 290 0},clip,width=0.98\linewidth,page=2]{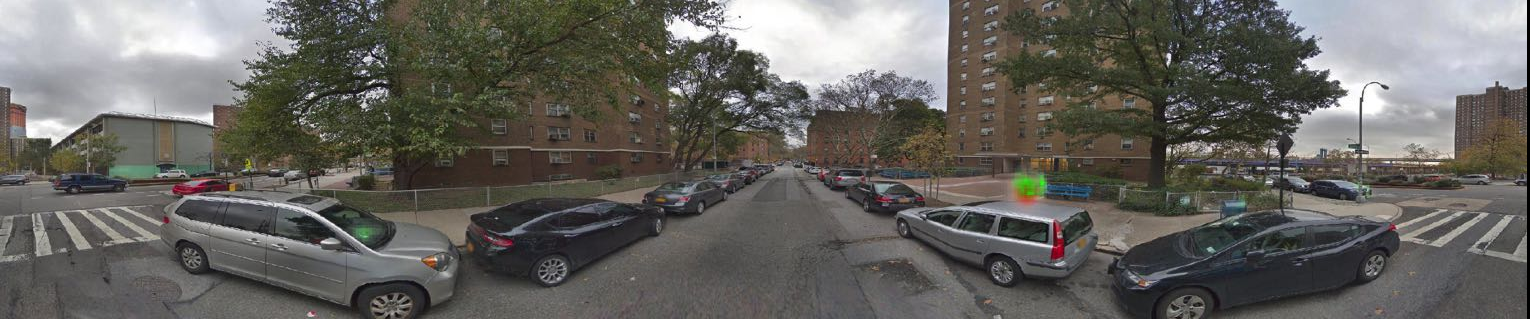}} \\[10pt]
\begin{minipage}{0.98\linewidth}
\noindent
\textbf{\texttoconv} The model predicts successfully as well. The distribution is focused on a smaller area compared to \lingunet closer to the top of the object. This possibly shows a learned bias towards placing Touchdown on the top of objects that \texttoconv is more suceptible to.
\end{minipage} 
\frame{\includegraphics[trim={0 445 290 170},clip,width=0.98\linewidth,page=2]{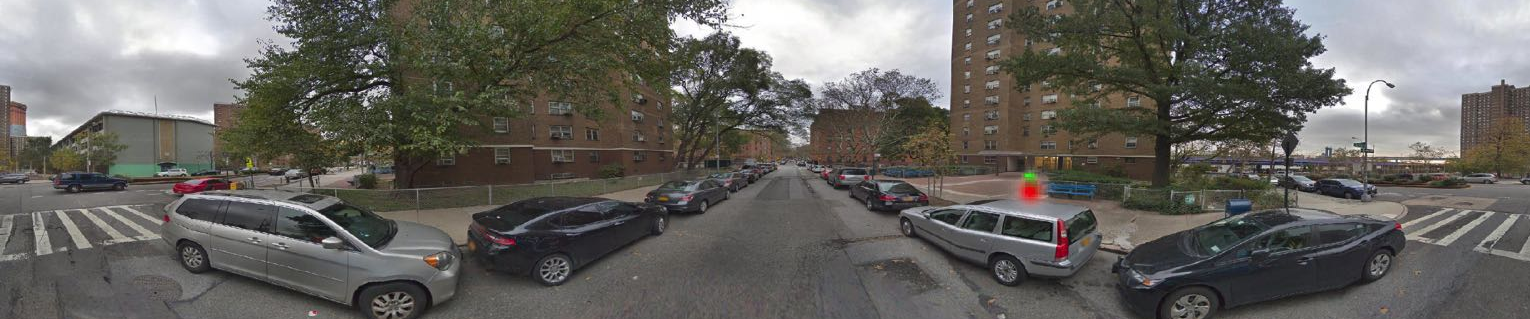}}\\[10pt]
\begin{minipage}{0.98\linewidth}
\noindent
\textbf{\imagetextconcatconv} The model prediction is correct.  The distribution is  focused on fewer pixels compared to \lingunet. 
\end{minipage} 
\frame{\includegraphics[trim={0 280 290 340},clip,width=0.98\linewidth,page=2]{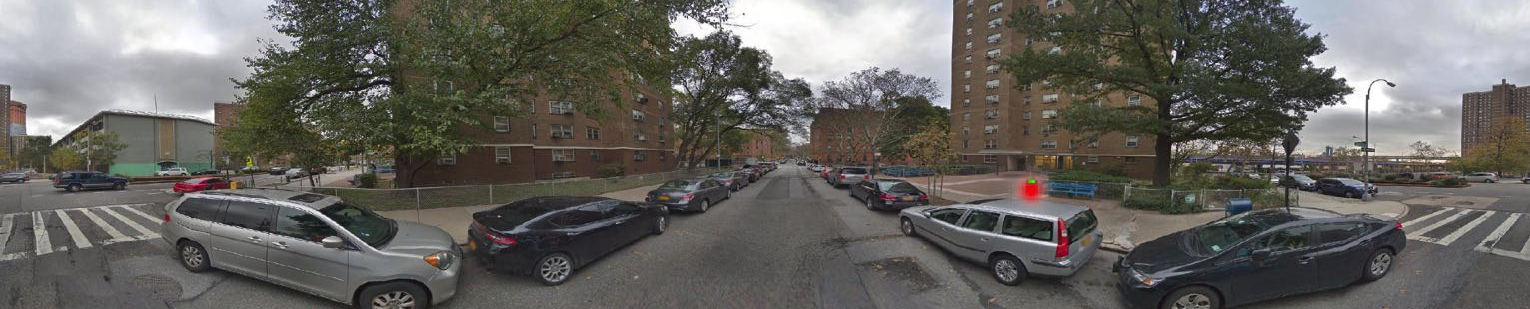}}\\[10pt]
\begin{minipage}{0.98\linewidth}
\noindent
\textbf{\imagetextconcat} The model prediction is correct. Similar to \imagetextconcatconv, it focuses on a few pixels.
\end{minipage} 
\frame{\includegraphics[trim={0 105 290 510},clip,width=0.98\linewidth,page=2]{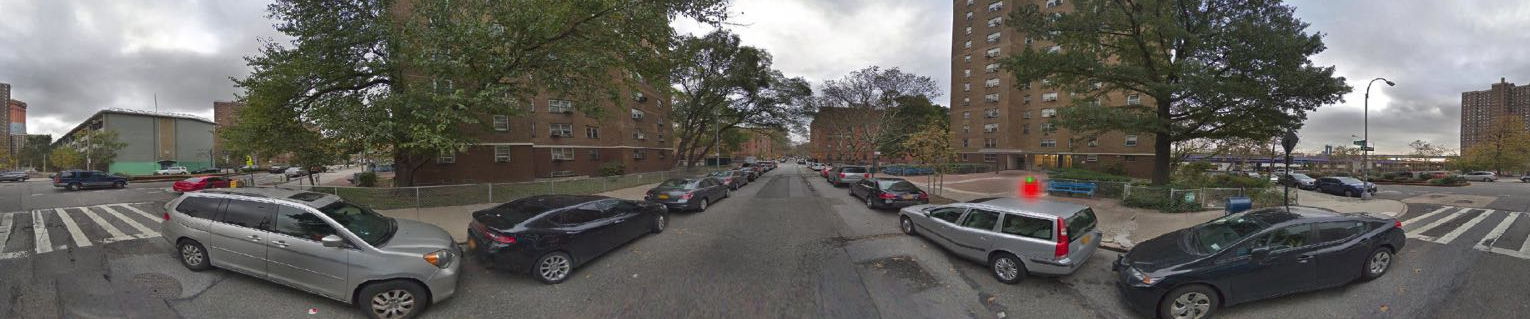}}
\caption{All the models predict the location of Touchdown correctly. Trash can is a relatively common object that workers use to place Touchdown in the dataset .}
\label{fig:sdr_sup_exp_2}
\end{figure*}

\begin{figure*}[t!]\centering
\fbox{\begin{minipage}{0.98\linewidth}
\nlstring{on your right is a parking garage, there is a red sign with bikes parked out in front of the garage, the bear is on the red sign.}
\end{minipage}} \\[10pt]
\begin{minipage}{0.98\linewidth}
\noindent
\textbf{\lingunet} The model predicted the location of Touchdown correctly to the red stop sign on the right side. 
\end{minipage} 
\frame{\includegraphics[trim={0 615 290 0},clip,width=0.98\linewidth,page=3]{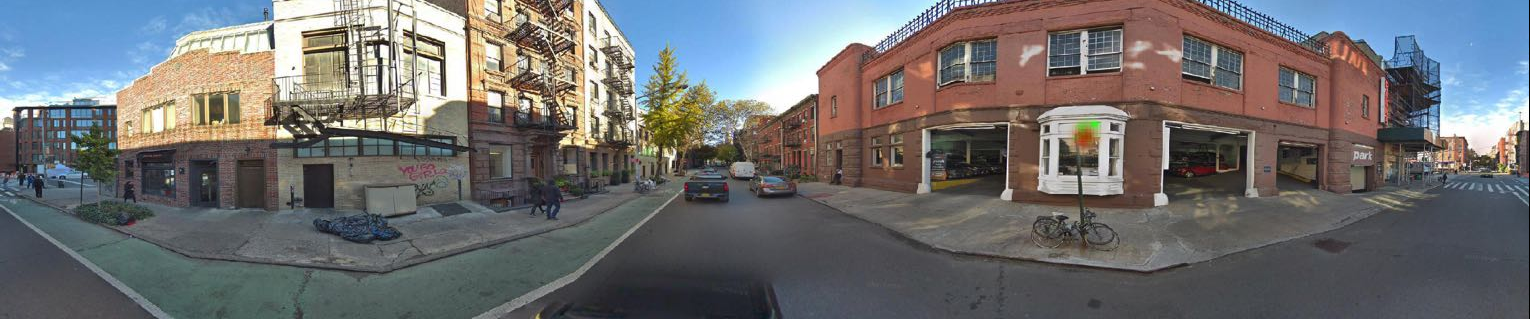}}\\[10pt]
\begin{minipage}{0.98\linewidth}
\noindent
\textbf{\texttoconv} The model predicts the location of Touchdown correctly. \end{minipage} 
\frame{\includegraphics[trim={0 445 290 170},clip,width=0.98\linewidth,page=3]{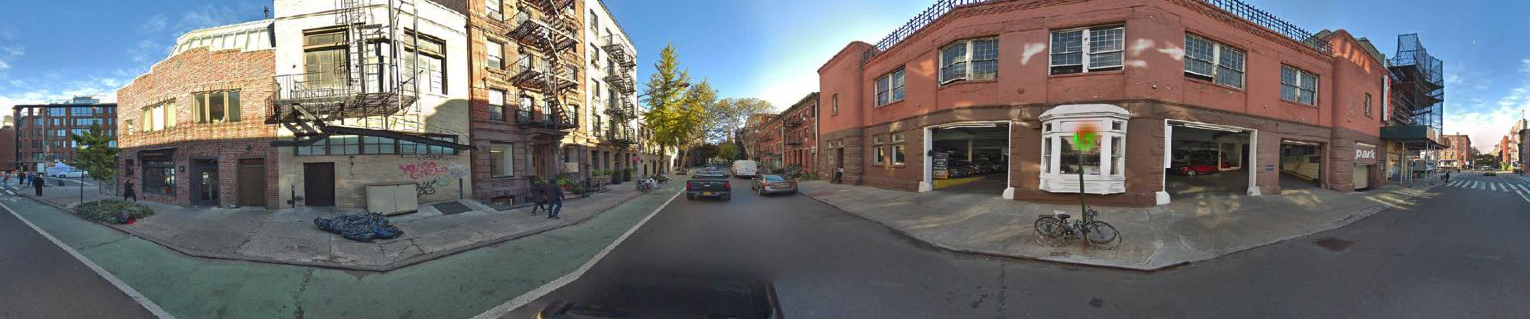}}\\[10pt]
\begin{minipage}{0.98\linewidth}
\noindent
\textbf{\imagetextconcatconv} The model predicts the location of Touchdown correctly. 
\end{minipage} 
\frame{\includegraphics[trim={0 280 290 340},clip,width=0.98\linewidth,page=3]{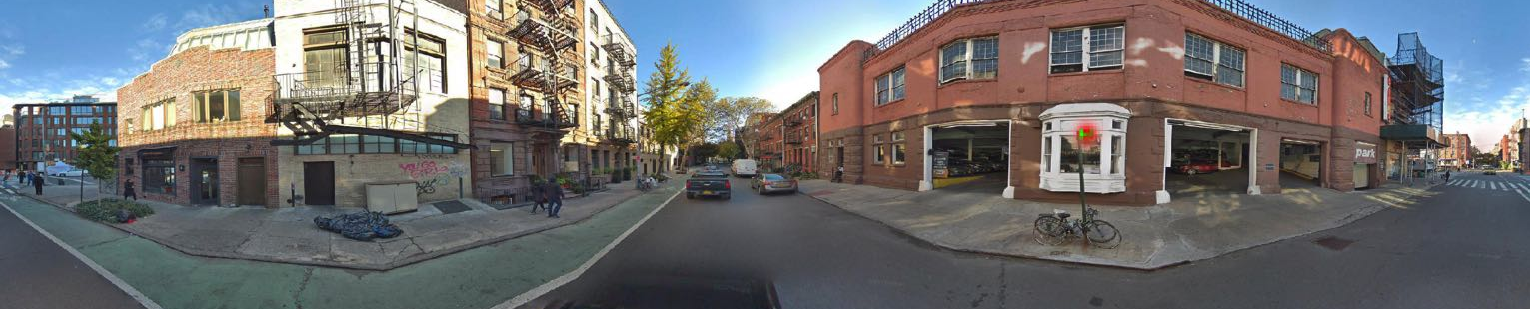}}\\[10pt]
\begin{minipage}{0.98\linewidth}
\noindent
\textbf{\imagetextconcat} The model predicts the location of Touchdown correctly. 
\end{minipage} 
\frame{\includegraphics[trim={0 105 290 510},clip,width=0.98\linewidth,page=3]{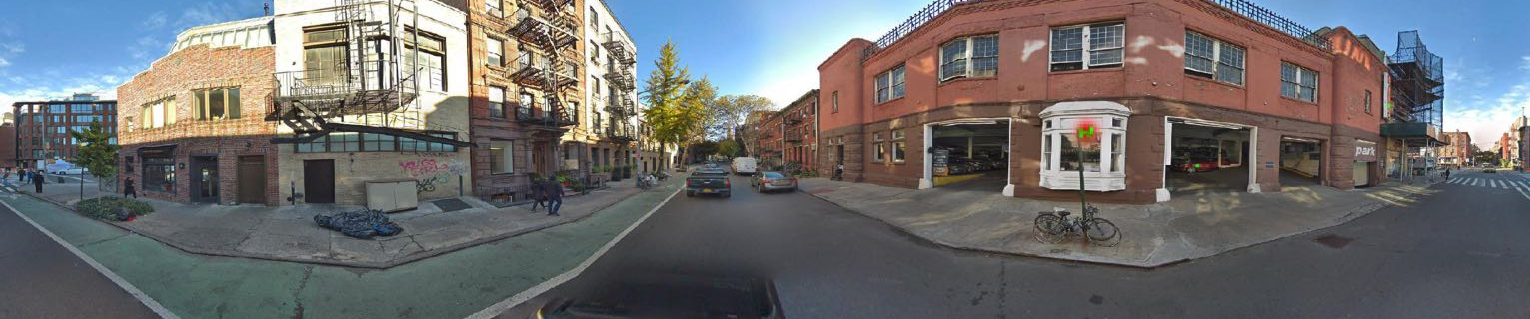}}
\caption{All the models predict the location of Touchdown correctly. Reference to a \nlstring{red sign} are relatively common in the data (Figure~\ref{fig:pie}) potentially simplifying this prediction.}
\label{fig:sdr_sup_exp_3}
\end{figure*}

\begin{figure*}[t!]\centering
\fbox{\begin{minipage}{0.98\linewidth}
\nlstring{touch down will be chillin in front of a sign on your right hand side about half way down this street,before you get to the sign there will be a multi color mural on the right w multiple colors and some writing on it.}
\end{minipage}} \\[10pt]
\begin{minipage}{0.98\linewidth}
\noindent
\textbf{\lingunet} The model fails to correctly predict the location of Touchdown, but is relatively close. The selected pixel is 104px from the correct one. The model focuses on the top of the sign instead of the bottom, potentially because of the more common reference to the top, which is visually distinguished. 
\end{minipage}
\frame{\includegraphics[trim={0 615 290 0},clip,width=0.98\linewidth,page=4]{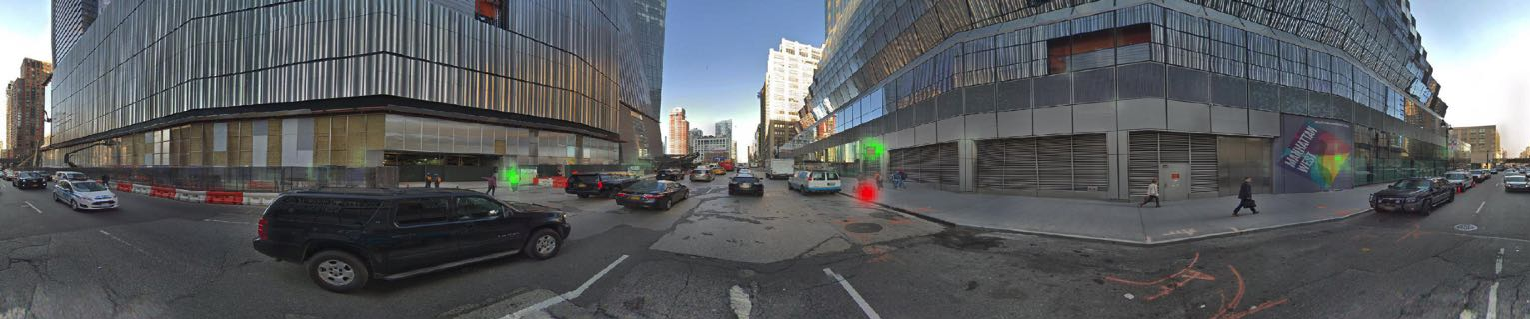}} \\[10pt]
\begin{minipage}{0.98\linewidth}
\noindent
\textbf{\texttoconv} The model fails to correctly predict the location of Touchdown, but is relatively close. The selected pixel is 96px from the correct one. 
\end{minipage}
\frame{\includegraphics[trim={0 445 290 170},clip,width=0.98\linewidth,page=4]{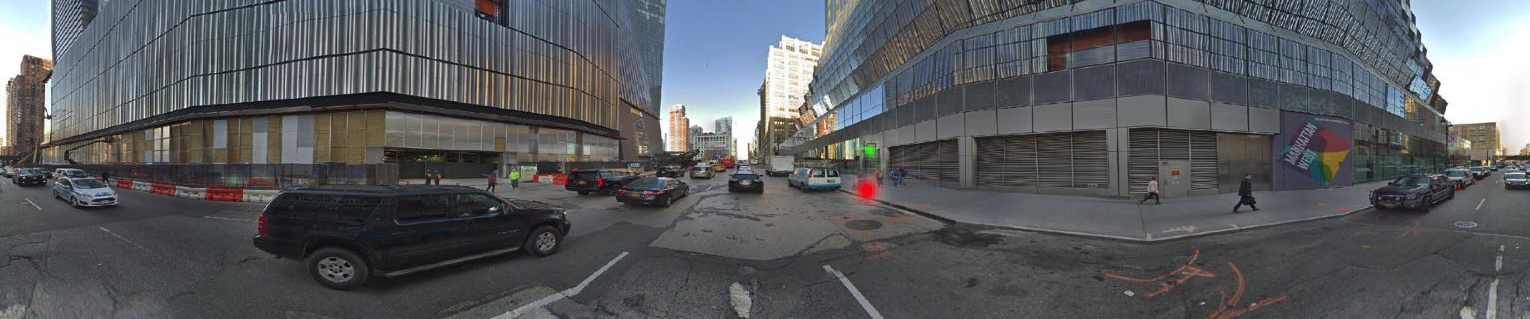}} \\[10pt]
\begin{minipage}{0.98\linewidth}
\noindent
\textbf{\imagetextconcatconv} The model fails to predict the location of Touchdown, instead focusing on a person walking on the left.
\end{minipage}
\frame{\includegraphics[trim={0 280 290 340},clip,width=0.98\linewidth,page=4]{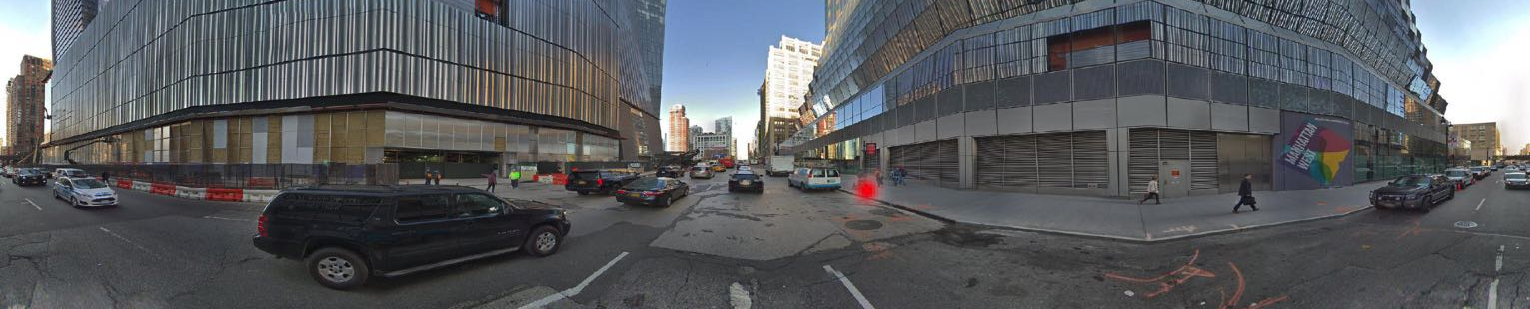}} \\[10pt]
\begin{minipage}{0.98\linewidth}
\noindent
\textbf{\imagetextconcat} The model fails to predict the location of Touchdown, instead of focusing the person walking on the left, the colorful sign mentioned in the description, and a car on the far right. 
\end{minipage}
\frame{\includegraphics[trim={0 105 290 510},clip,width=0.98\linewidth,page=4]{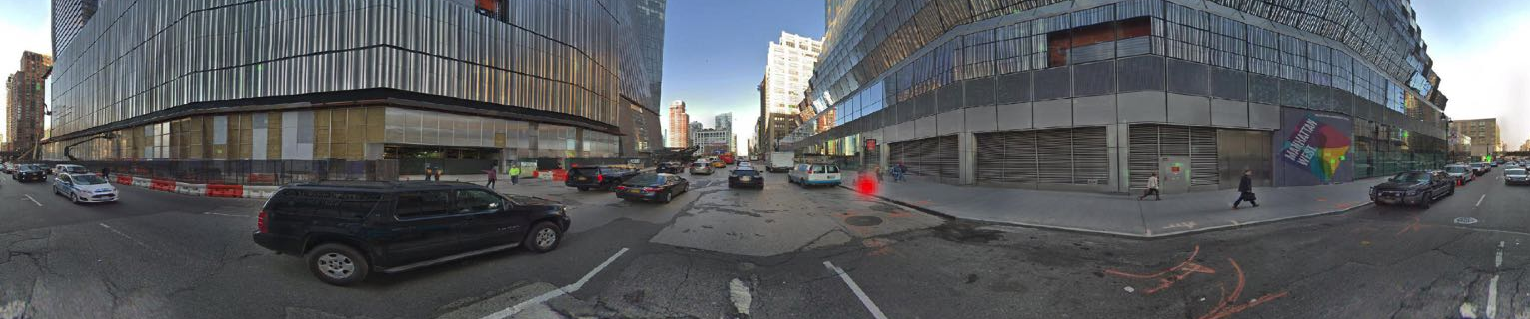}}
\caption{All the models fail to correctly identify the location of Touchdown. The predictions of \lingunet, \texttoconv, and \imagetextconcatconv seem to mix biases in the data with objects mentioned in the description, but fail to resolve the exact spatial description.}
\label{fig:sdr_sup_exp_4}
\end{figure*}

\begin{figure*}[t!]\centering
\fbox{\begin{minipage}{0.98\linewidth}
\nlstring{a row of blue bikes, touchdown is in the fifth bike seat in the row, from the way you came.}
\end{minipage}} \\[10pt]
\begin{minipage}{0.98\linewidth}
\noindent
\textbf{\lingunet} The model correctly identifies that a bike is mentioned, but fails to identify the exact bike or the location on the bike \nlstring{seat}. Instead the the distribution is divided between multiple bikes. 
\end{minipage}
\frame{\includegraphics[trim={0 615 290 0},clip,width=0.98\linewidth,page=5]{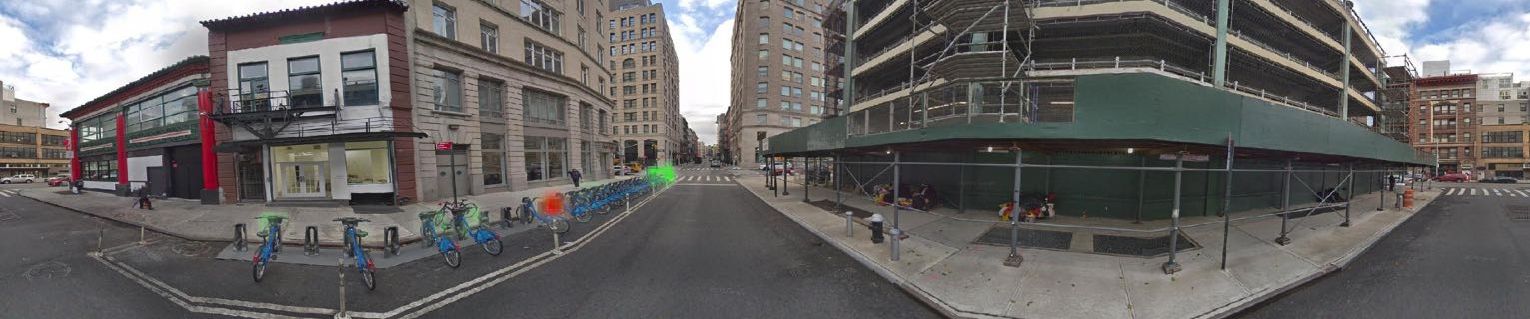}} \\[10pt]
\begin{minipage}{0.98\linewidth}
\noindent
\textbf{\texttoconv} Similar to \lingunet, the model identifies the reference to \nlstring{bikes}, but fails to identify the exact bike. The uncertainty of the model is potentially illustrated by how it distributes the probability mass. 
\end{minipage}
\frame{\includegraphics[trim={0 445 290 170},clip,width=0.98\linewidth,page=5]{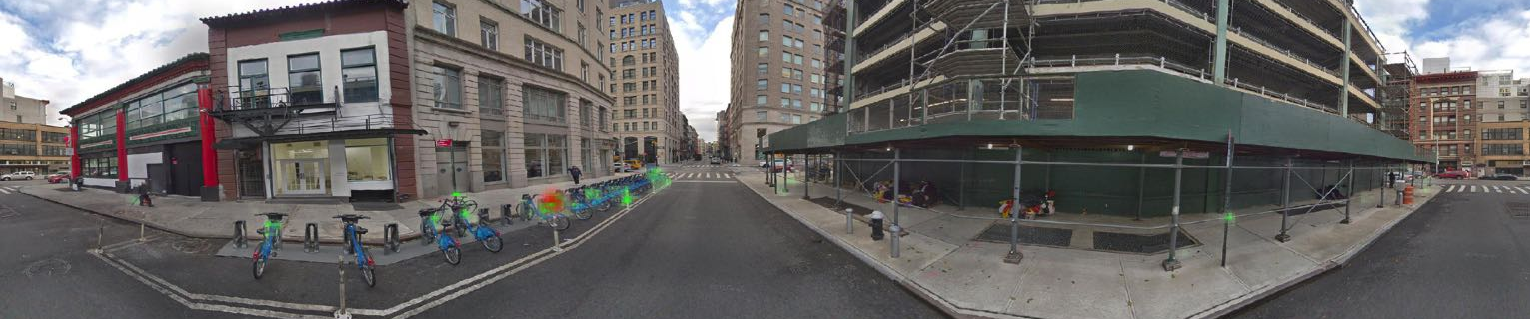}} \\[10pt]
\begin{minipage}{0.98\linewidth}
\noindent
\textbf{\imagetextconcatconv} The model correctly predicts the location of Touchdown. While the distribution is spread across multiple bikes observed, the highest probability pixel is close enough (i.e., within $80$ pixels) of the correct location. 
\end{minipage}
\frame{\includegraphics[trim={0 280 290 340},clip,width=0.98\linewidth,page=5]{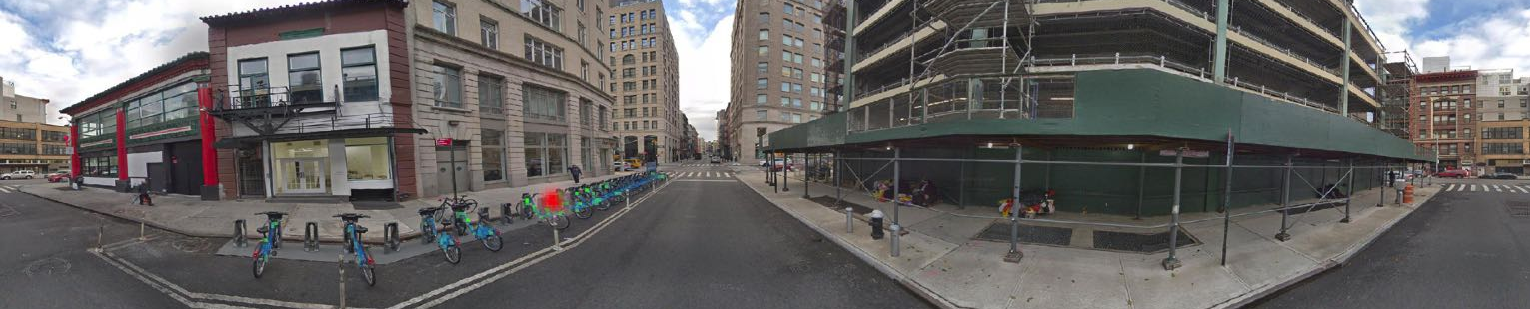}} \\[10pt]
\begin{minipage}{0.98\linewidth}
\noindent
\textbf{\imagetextconcat} Similar to \imagetextconcatconv, the model correctly predicts the location of Touchdown. 
\end{minipage}
\frame{\includegraphics[trim={0 105 290 510},clip,width=0.98\linewidth,page=5]{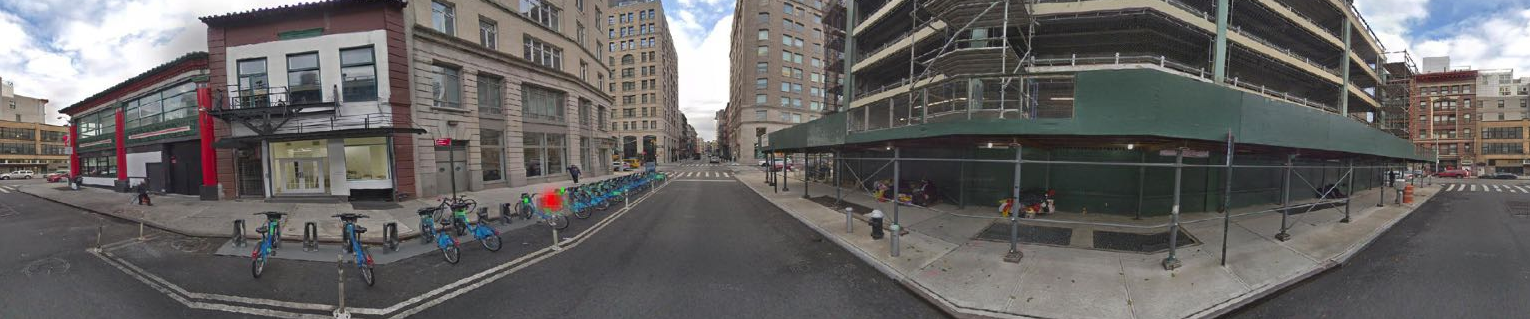}}
\caption{\lingunet and \texttoconv fail to correctly identify the location, although their predicted distribution is focused on the correct set of objects. In contrast, the simpler models, \imagetextconcat and \imagetextconcatconv, correctly predict the location of Touchdown. }
\label{fig:sdr_sup_exp_5}
\end{figure*}

\begin{figure*}[t!]\centering
\fbox{\begin{minipage}{0.98\linewidth}
\nlstring{on your right is a parking garage, there is a red sign with bikes parked out in front of the garage, the bear is on the red sign.}
\end{minipage}} \\[10pt]
\begin{minipage}{0.98\linewidth}
\noindent
\textbf{\lingunet} The model misidentifies the red sign on the left hand side as the correct answer. It fails to resolve the spatial description, instead focusing on a more salient \nlstring{red} object. 
\end{minipage}
\frame{\includegraphics[trim={0 615 290 0},clip,width=0.98\linewidth,page=6]{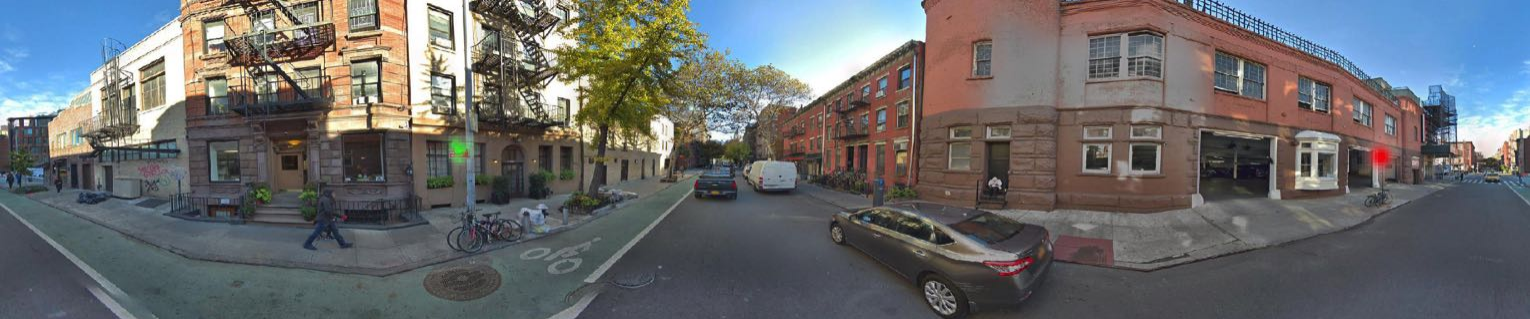}} \\[10pt]
\begin{minipage}{0.98\linewidth}
\noindent
\textbf{\texttoconv} The model fails to predict the correct location, instead focusing on the red sign closer to the center. 
\end{minipage}
\frame{\includegraphics[trim={0 445 290 170},clip,width=0.98\linewidth,page=6]{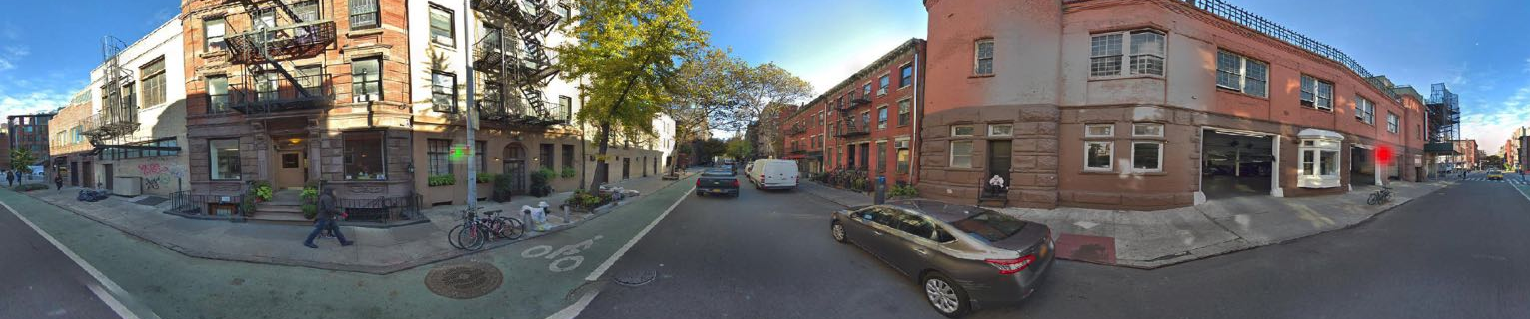}} \\[10pt]
\begin{minipage}{0.98\linewidth}
\noindent
\textbf{\imagetextconcatconv} The model fails to predict the correct location, instead focusing on the red sign closer to the center. 
\end{minipage}
\frame{\includegraphics[trim={0 280 290 340},clip,width=0.98\linewidth,page=6]{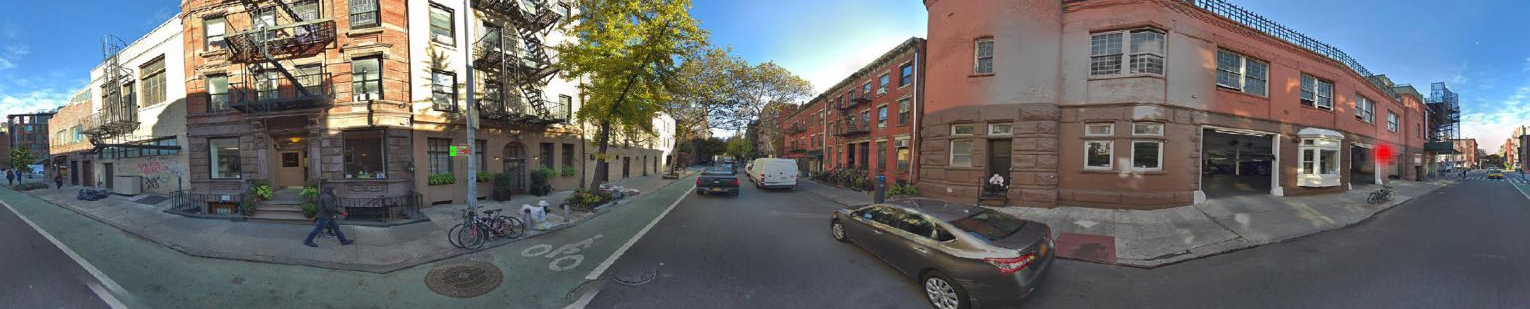}} \\[10pt]
\begin{minipage}{0.98\linewidth}
\noindent
\textbf{\imagetextconcat} The model fails to predict the correct location, instead focusing on the red sign close to the center of the image. 
\end{minipage}
\frame{\includegraphics[trim={0 105 290 510},clip,width=0.98\linewidth,page=6]{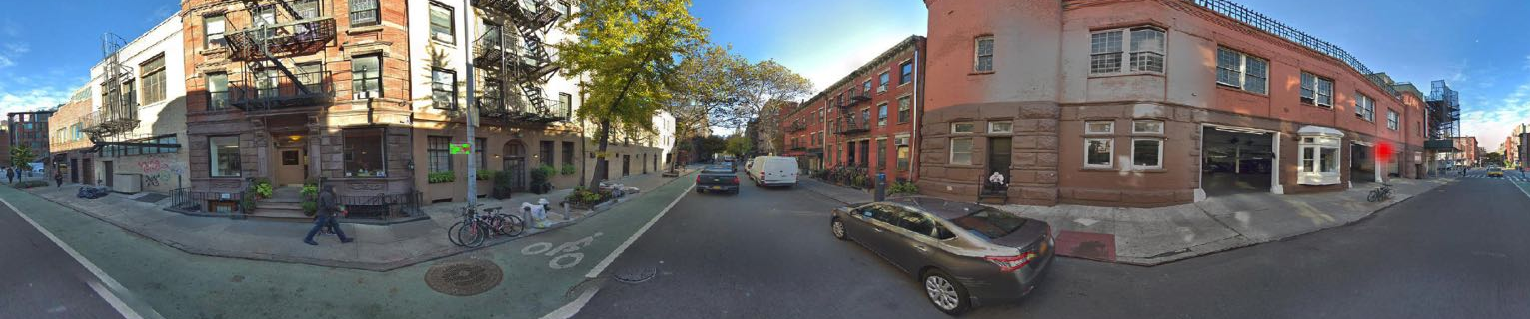}}
\caption{All the models fail to identify the correct location. They focus unanimously on the red sign on the left hand side. They all ignore the reference to the \nlstring{garage}, which is hard to resolve visually.}
\label{fig:sdr_sup_exp_6}
\end{figure*}

Figures~\ref{fig:sdr_sup_exp_1}--\ref{fig:sdr_sup_exp_6} show SDR pixel-level predictions for  comparing the four models we used:  {\lingunet}, {\imagetextconcat}, {\imagetextconcatconv}, and {\imagetextconcat}. 
Each figure shows the SDR description to resolve followed by the model outputs. 
We measure accuracy at a threshold of $80$ pixels. 
Red-overlaid pixels visualize the Gaussian smoothed annotated target location. Green-overlaid pixels visualize the model's probability distribution over pixels.

\section{SDR Experimental Setup Details}

\subsection{Models}

We use learned word vectors of size $300$.
For all models, we use a single-layer, bi-directional recurrent neural network (RNN) with long short-term memory (LSTM) cells~\cite{Hochreiter:97lstm} to encode the description into a fixed-size vector representation.
The hidden layer in the  RNN has $600$ unit. We compute the text embedding by averaging the RNN hidden states.

We provide the model with the complete panorama. 
We embed the panorama by  slicing it into eight images, and projecting each image  from a equirectangular projection to a perspective projection.
Each of the eight projected images is of size $800 \times 460$.
We pass each image separately through a \resnet~\cite{He:16resnet} pretrained on ImageNet~\cite{Russakovsky:15}, and extract features  from the fourth to last layer before classification; each slice's feature map is of size $128 \times 100 \times 58$.
Finally, the features for the eight image slices are concatenated into a single tensor of size $128 \times 100 \times 464$.

\para{\imagetextconcat} 
We concatenate the text representation along the channel dimension of the image feature map at each feature pixel and apply a multi-layer perceptron (MLP) over each pixel to obtain a real-value score for every pixel in the feature map. 
The multilayer perceptron includes two fully-connected layers with biases and $\rm ReLu$ non-linearities on the output of the first layer.
The hidden size of each layer is $128$.
A {\softmax} layer is applied to generate the final probability distribution over the feature pixels. 

\para{\imagetextconcatconv}
The network structure is the same as \imagetextconcat, except that after concatenating the text and image features and before applying the MLP, we mix the features across the feature map by applying a single convolution operation with a kernel of size $5 \times 5$ and padding of $2$. 
This operation does not change the size of the image and text tensor.
We use a the same MLP architecture as in \imagetextconcat on the outputs of the convolution, and compute a distribution over pixels with a {\softmax}.


\para{\texttoconv}
Given the text representation and the featurized image, we use a kernel conditioned on the text to convolve over the image.
The kernel is computed by projecting the text representation into a vector of size $409{,}600$ using a single learned layer without biases or non-linearities.
This vector is reshaped into a kernel of size $5 \times 5 \times 128 \times 128$, and used to convolve over the image features, producing a tensor of the same size as the featurized image.
We use a the same MLP architecture as in \imagetextconcat on the outputs of this operation, and compute a distribution over pixels  with a {\softmax}.

\para{\lingunet}
We apply two convolutional layers to the image features to compute $\mathbf{F}_1$ and $\mathbf{F}_2$.
Each uses a learned kernel of size $5 \times 5$ and padding of $2$. 
We split the text representation  into two vectors of size $300$, and use two separate learned layers  to transform each vector into another vector of size $16{,}384$ that is reshaped to $1 \times 1 \times 128 \times 128$.
The result of this operation on the first half of the text representation is $\mathbf{K}_1$, and on the second is $\mathbf{K}_2$.
The layers do not contain biases or non-linearities.
These two kernels are applied to $\mathbf{F}_1$ and $\mathbf{F}_2$ to compute $\mathbf{G}_1$ and $\mathbf{G}_2$.
Finally, we use two deconvolution operations in sequence on  $\mathbf{G}_1$ and $\mathbf{G}_2$ to compute $\mathbf{H}_1$ and $\mathbf{H_2}$ using learned kernels of size $5 \times 5$ and padding of $2$.

\subsection{Learning}
We initialize parameters by sampling uniformly from $[-0.1, 0.1]$.
During training, we apply dropout to the word embeddings with probability $0.5$.
We compute gradient updates using \textsc{Adam}~\cite{Kingma:14adam}, and  use a global learning rate of $0.0005$ for \lingunet, and $0.001$ for all other models.
We use early stopping with patience with  a  validation set containing $7\%$ of the training data to compute accuracy at a threshold of $80$ pixels after each epoch.
We begin with a patience of $4$, and when the accuracy on the validation set reaches a new maximum, patience resets to $4$.

\subsection{Evaluation} 
We compare the predicted location to the gold location by computing the location of the feature pixel corresponding to the gold location in the same scaling as the predicted probability distribution. We scale the accuracy threshold appropriately.

\subsection{\lingunet Architecture Clarifications}

Our \lingunet implementation for SDR task differs slightly from the original implementation~\cite{Blukis:18visit-predict}. We set the stride for both convolution and devonvolution operations to be $1$, whereas in the original \lingunet architecture the stride is set to $2$. 
Experiments with the original implementation show equivalent performance. 

\section{Navigation Experimental Setup Details}
\label{sup:nav-baseline}

\subsection{Models}

At each step, the agent observes the \emph{agent context}. 
Formally, the agent context $\tilde{s}$ at time step $t$ is a tuple $(\navinstruction, \panorama_t, \orientation_t, \langle (\panorama_1, \orientation_1, \action_1),\dots,(\panorama_{t-1}, \orientation_{t-1} \action_{t-1}) \rangle )$, where $\navinstruction$ is the navigation instruction, $\panorama_t$ is the panorama that is currently observed at heading $\orientation_t$, and $\langle (\panorama_1, \orientation_1, \action_1),\dots,(\panorama_{t-1}, \orientation_{t-1} \action_{t-1}) \rangle$ is the sequence of previously observed panoramas, orientations, and selected actions. 
Given an agent context $\tilde{s}$, the navigation model computes action probabilities  $P(\action \mid \tilde{s})$.

We use learned word vectors of size $32$ for all models. 
We map the instruction $\navinstruction$ to a vector $\mathbf{x}$ using a single-layer uni-directional RNN with LSTM cells with $256$ hidden units.
The instruction representation  $\mathbf{x}$ is the hidden state of the final token in the instruction.

We generate \resnet features for each $360^{\circ}$ panorama $\panorama_t$. We center the feature map according agent's heading $\orientation_t$. We crop a $128\times 100\times 100$ sized feature map from the center. We pre-compute mean value along the channel dimension for every feature map and save the resulting $100 \times 100$ features. 
This pre-computation allows for faster learning. 
We use the saved features corresponding to $\panorama_t$ and the agent's heading $\orientation_t$ as $\hat{\panorama}_t$.


\para{\concatmodel} 

We modify the model of  Mirowski \etal~\cite{mirowski2018learning} for instruction-driven navigation. 
We use an RNN to embed the instruction instead of a goal embedding, and  do not embed a reward signal. 
We apply a three-layer convolutional neural network to  $\hat{\panorama}_t$.
The first layer uses $32$ $8\times 8$ kernels with stride $4$, and the second layer uses $64$ $4\times 4$ kernels with stride $4$, applying $\rm ReLu$ non-linearities after each convolutional operation.
We use a single fully-connected layer including biases of size $256$ on the output of the convolutional operations to compute the observation's representation $\panorama'_t$.
We learn embeddings $\mathbf{a}$ of size $16$ for each action $\action$.
For each time step $t$, we concatenate the instruction representation $\mathbf{x}$, observation representation $\panorama'_t$, and action embedding $\mathbf{a}_{t-1}$ into a vector $\mathbf{\tilde{s}}_t$. For the first time step, we use a learned embedding for the previous action. We use a  single-layer RNN with $256$ LSTM cells on the sequence of time steps. The input at time $t$ is $\mathbf{\tilde{s}}_t$ and the hidden state is $\mathbf{h}_t$. 
We concatenate a learned time step embedding $\mathbf{t}\in\mathbb{R}^{32}$ with $\mathbf{h}_t$, and use a single-layer perceptron with biases and a $\softmax$ operation to compute $P(\action_t | \tilde{s}_t)$. 

\balance

\para{\gamodel} 
We apply a three-layer convolutional neural network to  $\hat{\panorama}_t$.
The first layer uses $128$ $8\times 8$ kernels with stride $4$, and the second layer uses $64$ $4\times 4$ kernels with stride $2$, applying $\rm ReLu$ non-linearities after each convolutional operation.
We use a single fully-connected layer including biases of size $64$ on the output of the convolutional operations to compute the observation's representation $\panorama'_t$.
We use a single hidden layer with biases followed by a sigmoid operation to map $\mathbf{x}$ into a vector $\mathbf{g} \in \mathbb{R}^{64}$. 
For each time step $t$, we apply a gated attention on $\panorama'_t$ using $\mathbf{g}$ along the channel dimension to generate a vector $\mathbf{u}_t$. 
We use  a single fully-connected layer with biases and a $\rm ReLu$ non-linearity with $\mathbf{u}_t$ to compute a vector $\mathbf{v}_t \in \mathbb{R}^{256}$.
We use a  single-layer RNN with $256$ LSTM cells on the sequence of time steps. The input at time $t$ is $\mathbf{v}_t$ and the hidden state is $\mathbf{h}_t$. 
We concatenate a learned time step embedding $\mathbf{t}\in\mathbb{R}^{32}$ with $\mathbf{h}_t$, and use a single-layer perceptron with biases and a $\softmax$ operation to compute $P(\action_t | \tilde{s}_t)$.


\subsection{Learning}

We train using asynchronous learning with  six clients, each using a different split of the training  data. We use supervised learning with \textsc{Hogwild!}~\cite{recht2011hogwild} and \textsc{Adam}~\cite{Kingma:14adam}. 
We  generate a sequence of agent contexts and actions $\{( \tilde{s}_i, \action_i)\}_{i=1}^N$ from the reference demonstrations, and  maximize the log-likelihood objective:

\begin{equation*}
    \mathcal{J} = \max_\theta \sum_{i=1}^N \ln p_\theta(\action_i \mid  \tilde{s}_i)\;\;,
\end{equation*}

\noindent
where $\theta$ is the model parameters.

\para{Hyperparameters}
We initialize parameters by sampling uniformly from $[-0.1,0.1]$.
We set the horizon to $55$ during learning, and use an horizon of $50$ during testing. 
We stop training using \spd performance on the development set. We use early stopping with patience, beginning  with a patience value of $5$ and resetting to $5$ every time we observe a new minimum \spd error. 
The global learning rate is fixed at $0.00025$.


%

\section{Additional Navigation Experiments}
\label{sec:app:nav}

\subsection{Experiments with RGB Images}
We also experiment with raw RGB images similar to Mirowski \etal~\cite{mirowski2018learning}. We project and resize each $360^{\circ}$ panorama $\panorama_t$  to a $60^{\circ}$ perspective image $\hat{\panorama}_t$ of size $3\times 84\times 84$, where the center of the panorama is the agent's heading  $\orientation_t$. Table~\ref{tab:rgb-results} shows the development and test results using RGB images. We observe better performance using \resnet features compared to RGB images.

\begin{table}[t!]
  \footnotesize
    \centering
    \begin{tabular}{|l|c|c|c|}
    \hline
    \multicolumn{1}{|c|}{\textbf{Method}} & \textbf{TC} & \textbf{SPD} & \textbf{SED} \\
    \hline\hline
    \multicolumn{4}{|l|}{\textbf{Development Results}} \\
    \hline\hline
        \concatmodel  & $\textbf{6.8}$ & $23.4$ & $\textbf{0.066}$\\
        \gamodel   & $6.5$ & $24.0$ & $0.064$ \\
        \hline\hline
            \multicolumn{4}{|l|}{\textbf{Test Results}} \\
    \hline\hline
        \concatmodel  & $\textbf{9.0}$  & $22.6$ & $\textbf{0.086}$\\
        \gamodel + \bcloning  & $7.9$ & $23.4$ & $0.076$\\
        \hline
    \end{tabular}
    \caption{Development and test navigation results using raw RGB images.}
    \label{tab:rgb-results}
\end{table}

\subsection{Single-modality Experiments}

We study the importance of each of the two modalities, language and vision,  for the navigation task. 
We separately remove the embeddings of the language (\textsc{NO-TEXT}) and visual observations (\textsc{NO-IMAGE}) for both models. 
Table~\ref{tab:modality} shows the results. 
We observe no meaningful learning in the absence of the vision modality, whereas limited  performance is possible without the natural language input. 
These results show both modalities are necessary, and also indicate that our navigation baselines (Table~\ref{tab:dev-nav-results}) benefit relatively little from the text.

\begin{table}[t!]
  \footnotesize
    \centering
    \begin{tabular}{|l|c|c|c|}
    \hline
    \multicolumn{1}{|c|}{\textbf{Method}} & \textbf{TC} & \textbf{SPD} & \textbf{SED} \\
    \hline\hline
    \multicolumn{4}{|l|}{\textbf{Development Results}} \\
    \hline\hline
     \concatmodel \textsc{NO-TEXT} & 24.48 & 7.26 & 0.07 \\
     \concatmodel \textsc{NO-IMAGE} & 35.68 & 0.22 & 0.001 \\
     \hline
     \gamodel \textsc{NO-TEXT} & 25.7 & 6.8 & 0.07\\
     \gamodel \textsc{NO-IMAGE} & 50.1 & 0.1 & 0.0 \\
     \hline
\end{tabular}
\caption{Single-modality development results.}
\label{tab:modality}
\end{table}

\clearpage

\clearpage

\clearpage

\clearpage

\clearpage

\clearpage

\clearpage

\end{document}